\documentclass[original]{clv2025}
\jvol{vv}
\jnum{nn}
\jyear{2026}

\dochead{Long Paper}

\runningtitle{Learning the Arabic Dialect Continuum}
\runningauthor{Khadraoui et al.}

\usepackage[utf8]{inputenc}
\usepackage{tabularx}
\usepackage{hyperref}
\usepackage{xcolor}
\usepackage{amsmath}
\usepackage{amssymb}
\usepackage{booktabs} 
\usepackage{array} 
\usepackage{caption}
\usepackage{enumitem}
\usepackage{graphicx}     
\usepackage{float}

\title{Learning the Arabic Dialect Continuum as a Continuous Space: A Regression Approach to Speaker Origin Prediction}

\author{
Mohamed Aziz Khadraoui$^{1}$,
Adel Ammar$^{2}$\thanks{Corresponding author: \email{aammar@psu.edu.sa}},
Bilel Benjdira$^{2}$,
Zahid Khan$^{2}$,
Skander Turki$^{2}$,
Wadii Boulila$^{2}$
}

\affilblock{
    \affil{Higher School of Communication of Tunis (SUP'COM), Tunisia}
    \affil{Robotics and Internet of Things Laboratory,
            Prince Sultan University,
            Riyadh, Saudi Arabia\\
            }
}

\begin{document}
\maketitle

\begin{abstract}
We present a regression-based approach to Arabic dialect geolocation that models dialectal variation as a continuous geographic space rather than discrete categories. Speaker origin is predicted as continuous latitude--longitude coordinates using a hierarchical neural architecture that fuses frame-level XLS-R-300M and Whisper-large-v3 encoder representations with phonotactic descriptors through a Transformer encoder and a learnable attention-pooled query. A spherical geodesic loss directly optimizes great-circle distance on Earth's surface, avoiding distortions inherent to planar coordinate regression. Under a leakage-free 5-fold GroupKFold protocol grouped by source recording, our model attains a pooled median localization error of 481.2\,km. Auxiliary country and city heads reach 64.5\% and 45.2\% accuracy, respectively. A permutation Mantel test on the learned latent space provides quantitative support for the Arabic dialect continuum hypothesis ($r = 0.131$, $p \approx 10^{-4}$ over 10{,}000 permutations). To probe true generalization, we further introduce a city-masking protocol in which two cities per fold are removed from training but retained in validation; under this zero-shot regime mean error rises to 1173.3\,km, a $1.32\times$ degradation relative to seen cities. Our findings establish continuous geographic modeling as a principled framework for Arabic dialect geolocation and quantify, both its strengths and the substantial headroom that remains.
\end{abstract}

\section{Introduction}

Arabic dialectal variation spans a vast geographic region from Morocco to Oman, encompassing hundreds of millions of speakers across 22 countries. While Modern Standard Arabic (MSA) serves as a formal written standard, spoken communication predominantly occurs in regional dialects that differ substantially in phonology, morphology, lexicon, and syntax. These differences are not merely surface-level: mutual intelligibility between geographically distant varieties (for example, Moroccan Darija and Gulf Arabic) can be severely limited, while neighbouring varieties shade into one another almost imperceptibly. Understanding and automatically identifying these dialectal differences has profound implications for speech recognition, machine translation, information retrieval, voice assistants, forensic speaker profiling, and sociolinguistic research.

The spoken modality is central to this problem. Dialectal distinctions are realized most richly in speech, through phonetic, prosodic, and rhythmic cues that are largely lost when utterances are transcribed into a non-standardized orthography. Consequently, while a large body of work has addressed dialect identification from \emph{text}, identifying dialect (and, more ambitiously, speaker origin) directly from the \emph{audio signal} is both more faithful to how dialects are used and considerably more challenging, owing to channel variability, background noise, code-switching with MSA or other languages, and the scarcity of geographically labelled speech corpora. Early speech-based systems framed the problem as a small, coarse classification task (typically distinguishing four or five regional groups) and only recently has city-level granularity become feasible with the release of large, metadata-rich speech corpora.

Traditional approaches to Arabic dialect identification, whether text- or speech-based, treat dialects as discrete, mutually exclusive categories corresponding to countries, regions, or cities. However, this discretization is in tension with established sociolinguistic theory, which characterizes dialectal variation as existing along continuous gradients (\emph{dialect continua}) rather than exhibiting sharp boundaries~\cite{chambers1998dialectology,heeringa2001dialect,nerbonne2009data}. Speakers near geographic borders often exhibit mixed dialectal features, and migration patterns create complex, overlapping isoglosses that defy simple categorization. Imposing hard categorical boundaries on this gradient structure discards precisely the information that distinguishes a principled geographic model from an arbitrary partition of space.

We propose a paradigm shift from discrete classification to continuous regression, formulating dialect geolocation as the prediction of precise geographic coordinates on Earth's surface. Conceptually, this reframes Arabic dialect identification as a speech-based geolocation problem, analogous to image geolocation in computer vision, where the goal is to place an input on a map rather than into a bin. This regression-based approach offers several advantages: (1) it naturally captures gradient variation without imposing artificial decision boundaries, (2) it enables fine-grained predictions at unseen locations, (3) it reduces sensitivity to class imbalance arising from uneven data collection, and (4) it provides a principled framework for modeling the spatial structure of linguistic variation. Crucially, it also yields an interpretable, distance-based error metric (kilometers on the Earth's surface) that is directly comparable across studies, unlike accuracy figures that depend on an arbitrary choice of dialect classes.

The main contributions of this work are as follows:
\begin{itemize}
    \item We formulate Arabic dialect geolocation as a \emph{continuous geographic regression} task, replacing conventional discrete dialect classification with the prediction of latitude-longitude coordinates on the Earth's surface.
    
    \item We propose a hierarchical multi-task architecture that combines spherical geodesic regression with auxiliary country and city classification objectives, enabling the model to jointly capture continuous and discrete aspects of dialectal variation.
    
    \item We introduce a rigorous evaluation protocol based on the ARCADE corpus~\cite{nacar2026arcade}, including geographically informed cross-validation with masked-city splits and geodesic distance metrics, providing a reproducible benchmark for continuous dialect geolocation.
    
    \item We provide an extensive empirical and linguistic analysis demonstrating that Arabic dialect variation is more accurately modeled as a continuous geographic continuum than as a collection of isolated dialect classes, yielding new insights into the relationship between geographic proximity and speech variation.
\end{itemize}

Experimental results on 2{,}329 Arabic speech samples spanning 19 countries and 46 cities, evaluated under a leakage-free 5-fold GroupKFold protocol with city-masking for zero-shot probing, demonstrate a pooled median localization error of 481.2\,km and country accuracy of 64.5\%. Beyond raw accuracy, we analyze model calibration, selective prediction, border-versus-interior behavior, and unseen-city generalization, establishing both practical pathways for deployment and the gap that remains between closed-set evaluation and true geographic generalization.

The remainder of this paper is organized as follows. Section~\ref{sec2} reviews the relevant literature on Arabic dialect identification, continuous dialect modeling, and speech-based geolocation. Section~\ref{sec3} formulates the task as a continuous geographic regression problem. Section~\ref{sec4} describes the dataset, data-quality procedures, and preprocessing pipeline. Section~\ref{sec5} presents the proposed methodology, including the model architecture, training objectives, and optimization strategy. Section~\ref{sec6} details the experimental setup and evaluation protocol, while Section~\ref{sec7} presents and discusses the experimental results. Section~\ref{sec8} discusses the limitations of the proposed approach. Finally, Section~\ref{sec9} concludes the paper and outlines directions for future research.

\section{Background and Related Works}\label{sec2}

\subsection{Early Dialect Identification Systems}

Early Arabic dialect identification systems treated dialects as discrete categories. Zaidan et al.~\cite{zaidan2014arabic} established the foundation with the Arabic Online Commentary dataset containing 52 million words, while Bouamor et al.~\cite{bouamor2018madar} created the comprehensive MADAR corpus with parallel translations across 25 city dialects. Traditional approaches using Support Vector Machines and n-gram features~\cite{elfardy2013sentence} achieved reasonable performance but failed to capture the continuous nature of dialectal variation. We note that all of these foundational systems operate on \emph{text}; the spoken modality, which is the focus of our work, has historically received far less attention.

\subsection{Speech-Based Dialect Identification}

In contrast to the large text-based literature, spoken Arabic dialect identification (ADI) has matured more slowly, gated by the availability of labelled speech. Ali et al.~\cite{ali2016automatic} introduced automatic dialect detection from Arabic broadcast speech, combining acoustic and lexical (phone- and word-level) features over five regional varieties, and Khurana et al.~\cite{khurana2016multiview} explored multi-view dimensionality reduction for the same task. The ADI17 benchmark of Shon et al.~\cite{shon2020adi17} scaled the problem to 17 country-level dialects using $\sim$3{,}000 hours of YouTube speech, and end-to-end neural systems built on i-vectors, x-vectors, and convolutional encoders became the dominant paradigm. More recently, self-supervised speech encoders such as wav2vec~2.0 and its multilingual variant XLS-R~\cite{babu2022xlsr}, together with weakly supervised models such as Whisper~\cite{radford2023whisper}, have provided strong transferable representations that we exploit in this work. Parallel efforts have expanded the available labelled speech: the Casablanca corpus~\cite{talafha2024casablanca}, for instance, provides multi-dialectal Arabic speech with transcriptions and dialect labels across several countries. Critically, however, all of these speech systems remain \emph{classification} systems over a small, fixed set of regions or countries; none predicts continuous speaker origin, and city-level granularity has only become tractable with the recent release of the metadata-rich ARCADE radio corpus~\cite{nacar2026arcade} on which we build.

\subsection{Continuous Dialect Modeling}

The revolutionary shift came with Keleg et al.~\cite{keleg2023aldi}, who introduced regression-based modeling that quantifies dialectness continuously. Their AOC-ALDi dataset contains 127,835 sentences with continuous annotations, achieving RMSE values of 0.1403. Building on this, Shaban et al.~\cite{shaban2024ags} added the Arabic Generality Score dimension, employing etymology-aware edit distance and CAMeL-BERT regression. This two-dimensional continuous space (ALDi $\times$ AGS) provides unprecedented granularity in modeling dialectal relationships.

\subsection{Transformer-Based Architectures}

The transformer revolution has profoundly impacted Arabic dialect processing. AraBERT~\cite{antoun2020arabert} pioneered transformer-based Arabic understanding, while MARBERTv2~\cite{abdul2020marbert}, trained on 6 billion tweets, achieves 65\% accuracy on 18-dialect classification. Recent advances in the NADI shared tasks~\cite{abdul2024nadi} demonstrate the effectiveness of ensemble regression methods combining BERT-based models with contrastive learning. Elaraby et al.~\cite{elaraby2018deep} showed that BiLSTM architectures achieve 92.65\% accuracy, while parameter-efficient approaches demonstrate comparable performance using only 2.5\% of model parameters.

\subsection{Key Datasets and Resources}

The MADAR corpus~\cite{bouamor2018madar} remains the most comprehensive resource with 47,466 dialectal words mapped to 1,045 concepts across 25 city dialects. QADI~\cite{abdelali2021qadi} contributes 540,590 tweets from 18 Arab countries with 91.5\% intrinsic accuracy. The Arabic Online Commentary dataset~\cite{zaidan2011aoc} provides foundational coverage of Egyptian, Gulf, and Levantine dialects. For continuous modeling, the AOC-ALDi corpus with continuous dialectness annotations and the NADI 2024 dataset~\cite{abdul2024nadi} featuring multi-label geographic annotations are essential. These resources enable fine-grained analysis at unprecedented granularity, supporting regression-based approaches to speaker origin prediction.

\subsection{Challenges and Open Problems}

Code-switching between Modern Standard Arabic and dialects presents fundamental challenges, with word error rates of 28--54\% for code-switched speech~\cite{habibi2025codeswitching}. The lack of standardized orthography, addressed partially by CODA~\cite{habash2012coda}, continues to impact model generalization. Data scarcity affects numerous dialects disproportionately. In fact, while Egyptian Arabic has extensive resources, rural North African variants remain severely underrepresented. Evaluation metrics for continuous predictions remain inadequate. Traditional metrics like accuracy fail to capture the nuanced nature of continuous dialectness, while regression metrics like RMSE don't fully represent linguistic validity. The absence of standardized evaluation frameworks hinders reproducible research.

\subsection{Tools and Frameworks}

The tools most widely used for Arabic dialect work operate on \emph{text} rather than speech. CAMeL Tools~\cite{obeid2020camel} provides comprehensive Arabic \emph{text} NLP capabilities including text-based dialect identification for 25 dialects. MADAMIRA~\cite{pasha2014madamira} offers sophisticated \emph{textual} morphological analysis and disambiguation. AraVec~\cite{soliman2017aravec} provides pre-trained \emph{word} embeddings learned from written corpora. These text-oriented resources underpin most modern Arabic dialect systems but are not directly applicable to the raw audio signal; speech-based systems instead rely on acoustic toolkits (e.g.\ Kaldi-style i-vector/x-vector pipelines) and, increasingly, on self-supervised speech encoders such as XLS-R~\cite{babu2022xlsr} and Whisper~\cite{radford2023whisper}, as adopted in this work.

\subsection{Comparison with Prior Work}

Table~\ref{tab:related_comparison} situates our approach against representative prior work along the dimensions most relevant to this paper: the input modality (text vs.\ speech), the task formulation, the geographic granularity of the labels, and the form of the output. The table makes explicit that prior speech-based systems are classifiers over a handful of regions or countries, prior continuous-modeling work (ALDi/AGS) operates on text and predicts a dialectness score rather than a location, and our work is the only entry that predicts continuous geographic coordinates directly from speech.

% \begin{table}[htbp]
% \centering
% \caption{Comparison of representative Arabic dialect identification work. ``Granularity'' refers to the geographic resolution of the labels; ``Output'' to what the system predicts. Our work is the only one combining the \emph{speech} modality with \emph{continuous} geographic regression.}
% \label{tab:related_comparison}
% \small
% \begin{tabular}{l c c c l}
% \toprule
% \textbf{Work} & \textbf{Modality} & \textbf{Task} & \textbf{Granularity} & \textbf{Output} \\
% \midrule
% Zaidan \& Callison-Burch~\cite{zaidan2014arabic} & Text   & Classification & 4 regions        & Dialect label \\
% Bouamor et al.\ (MADAR)~\cite{bouamor2018madar}   & Text   & Classification & 25 cities        & City label \\
% Abdul-Mageed et al.\ (NADI)~\cite{abdul2024nadi}  & Text   & Classification & 18--21 countries & Country/province label \\
% Keleg et al.\ (ALDi)~\cite{keleg2023aldi}         & Text   & Regression     & ---              & Dialectness score $\in[0,1]$ \\
% Ali et al.~\cite{ali2016automatic}                & Speech & Classification & 5 regions        & Dialect label \\
% Shon et al.\ (ADI17)~\cite{shon2020adi17}         & Speech & Classification & 17 countries     & Country label \\
% Nacar et al.\ (ARCADE)~\cite{nacar2026arcade}     & Speech & Classification & 58 cities        & City/dialect tag \\
% \textbf{This work}                                & \textbf{Speech} & \textbf{Regression} & \textbf{Coordinates} & \textbf{(lat, lon) on the sphere} \\
% \bottomrule
% \end{tabular}
% \end{table}

\begin{table}[htbp]
\centering
\caption{Comparison of representative Arabic dialect identification work. ``Granularity'' refers to the geographic resolution of the labels; ``Output'' to what the system predicts. Our work is the only one combining the \emph{speech} modality with \emph{continuous} geographic regression.}
\label{tab:related_comparison}
\footnotesize
\setlength{\tabcolsep}{4pt}
\begin{tabularx}{\linewidth}{@{}l c c c >{\raggedright\arraybackslash}X@{}}
\toprule
\textbf{Work} & \textbf{Modality} & \textbf{Task} & \textbf{Granularity} & \textbf{Output} \\
\midrule
\cite{zaidan2014arabic}
& Text & Classification & 4 regions & Dialect label \\

\cite{bouamor2018madar}
& Text & Classification & 25 cities & City label \\

\cite{abdul2024nadi}
& Text & Classification & 18--21 countries & Country/province label \\

ALDi~\cite{keleg2023aldi}
& Text & Regression & --- & Dialectness score $\in[0,1]$ \\

\cite{ali2016automatic}
& Speech & Classification & 5 regions & Dialect label \\

ADI17~\cite{shon2020adi17}
& Speech & Classification & 17 countries & Country label \\

ARCADE~\cite{nacar2026arcade}
& Speech & Classification & 58 cities & City/dialect tag \\

\textbf{This work}
& \textbf{Speech} & \textbf{Regression} & \textbf{Coordinates}
& \textbf{(lat, lon) on the sphere} \\
\bottomrule
\end{tabularx}
\end{table}

\subsection{Motivation for This Work}

Despite significant progress, existing work has two critical limitations. First, most systems continue to frame dialect identification as discrete classification, imposing categorical boundaries that contradict sociolinguistic evidence of continuous variation. Second, geographic-linguistic modeling incorporating spatial relationships between dialect communities remains underexplored for speaker origin prediction. Our work addresses these gaps by developing a continuous regression framework grounded in spherical geometry, providing both improved localization accuracy and theoretical insights into the structure of Arabic dialectal variation.

\section{Problem Formulation}\label{sec3}

In this section, we formally define the task of fine-grained Arabic dialect geolocation. Unlike traditional approaches that treat dialect identification as a discrete classification problem over broad and often arbitrary regions, we frame the objective as a continuous geographic regression task aimed at directly predicting a speaker's spatial coordinates. To robustly model this continuous space, we map standard latitude and longitude representations into a 3D spherical coordinate system, mitigating inherent planar distortions and boundary discontinuities. Finally, recognizing that human geography is intrinsically structured, we establish a hierarchical multi-task learning framework. This framework jointly optimizes continuous coordinate regression alongside categorical country and city classification, thereby enriching the supervisory signal and ensuring geographically consistent predictions.

\subsection{Dialect Recognition as a Continuous Geographic Regression Task}

We formulate Arabic dialect geolocation as predicting continuous geographic coordinates from speech input. Given an audio signal $\mathbf{x}$, the task is to estimate the speaker's origin as latitude-longitude coordinates $(\phi, \lambda)$ where $\phi \in [-90^\circ, 90^\circ]$ and $\lambda \in [-180^\circ, 180^\circ]$.

Traditional dialect classification approaches discretize this space into $K$ predefined regions $\{R_1, \ldots, R_K\}$ and optimize:
\begin{equation}
\hat{k} = \arg\max_{k \in \{1,\ldots,K\}} P(R_k \mid \mathbf{x}).
\end{equation}
The predicted location is then the centroid of region $R_{\hat{k}}$. This discretization introduces several problems: (1) artificial boundaries create discontinuities in prediction, (2) within-region variation is ignored, (3) predictions at unseen locations require defining new categories, and (4) class imbalance from uneven geographic sampling creates optimization difficulties.

In contrast, our regression approach directly predicts geographic coordinates:
\begin{equation}
(\hat{\phi}, \hat{\lambda}) = f_\theta(\mathbf{x}),
\end{equation}
where $f_\theta$ is a neural network with parameters $\theta$. This formulation naturally handles gradient variation, enables interpolation to unseen locations, and provides fine-grained localization without categorical constraints.

\subsection{Spherical Geometry and Coordinate Representation}

Operating directly on latitude-longitude pairs introduces several challenges. First, longitude is periodic with a discontinuity at $\pm 180^\circ$. Second, distance metrics in $(\phi, \lambda)$ space do not correspond to actual geographic distance: one degree of longitude at the equator spans $\sim$111 km, while at $60^\circ$ latitude it spans only $\sim$56 km. Third, mean squared error in coordinate space distorts spatial relationships near the poles. While these issues are less critical when the scope is restricted solely to the Arab world region, given its limited latitude and longitude ranges, we aim for a generalized approach that remains robust across broader geographic contexts.

We address these issues by representing locations as unit vectors on the sphere. Each coordinate pair $(\phi, \lambda)$ is mapped to 3D Cartesian coordinates:
\begin{equation}
\left\{
\begin{aligned}
x &= \cos(\phi) \cos(\lambda) \\
y &= \cos(\phi) \sin(\lambda) \\
z &= \sin(\phi)
\end{aligned}
\right.
\end{equation}
yielding a unit vector $\mathbf{y} = (x, y, z)$ with $\|\mathbf{y}\| = 1$. In this representation, the great-circle distance between two points is simply:
\begin{equation}
d(\mathbf{y}_1, \mathbf{y}_2) = R \cdot \arccos(\mathbf{y}_1 \cdot \mathbf{y}_2),
\end{equation}
where $R = 6371$ km is Earth's mean radius. This formulation eliminates discontinuities, ensures distance computation along Earth's surface, and simplifies loss function design.

\subsection{Hierarchical Geographic Structure}

Geographic variation exhibits natural hierarchy: countries contain regions, regions contain cities, and cities contain neighborhoods. We exploit this structure through multi-task learning, jointly predicting:
\begin{itemize}[leftmargin=*, nosep]
\item Fine-grained coordinates $(\phi, \lambda)$ via regression.
\item Country label $c \in \{1, \ldots, C\}$ via classification.  
\item City label $s \in \{1, \ldots, S\}$ via classification.
\end{itemize}

The hierarchical tasks serve multiple purposes: (1) auxiliary classification objectives provide additional supervision signal, (2) predicted country/city labels can condition coordinate prediction, improving accuracy, (3) classification confidence scores enable uncertainty quantification, and (4) hierarchical predictions offer graceful degradation when fine-grained localization fails.

\section{Dataset}\label{sec4}

\subsection{Data Sources}

Our work builds on the ARCADE corpus~\cite{nacar2026arcade}, a city-scale collection of Arabic radio speech segmented into short clips, each annotated with metadata including the broadcasting station, its city, and its country; we refer the reader to the ARCADE paper for the full description of the streaming-based collection, segmentation, and annotation pipeline. From this collection we derive our working set through the data-quality filtering described below. After filtering, our dataset comprises 2{,}329 radio recordings across 19 countries and 46 cities. The retained recordings span dialectal regions including Maghrebi, Egyptian, Levantine, Iraqi, Gulf, Sudanese, and Yemeni varieties. For each recording we attach a precise latitude--longitude coordinate obtained by geocoding the (city,~country) pair against the Nominatim geocoding service~\cite{nominatim}.

\subsection{Data Collection and Quality Control}

The audio was not collected by us through elicitation; it consists of naturalistic radio-broadcast speech harvested from streaming services across the Arab world, as detailed in the ARCADE paper~\cite{nacar2026arcade}. Our contribution at the data level is the quality filtering applied on top of that collection to produce a clean, geographically consistent subset suitable for coordinate regression. Concretely, starting from the ARCADE clips we (1) discard cities with fewer than five samples (folded into an \texttt{\_\_OTHER\_\_} bucket during training, see Section~\ref{sec:cv-protocol}), (2) remove geographic outliers whose geocoded coordinates fall more than $2000$\,km from their country centroid (typically the result of ambiguous or mis-resolved city names), and (3) drop clips with missing or unresolvable city/country metadata. This filtering reduces the raw collection to the 2{,}329 recordings used throughout this paper. We deliberately do not re-describe the upstream recording-quality and dialect-validity checks, which are the responsibility of the ARCADE pipeline and are documented in~\cite{nacar2026arcade}.

\subsection{Geographic Distribution}

\begin{figure}[htbp]
    \centering
    \includegraphics[width=\textwidth]{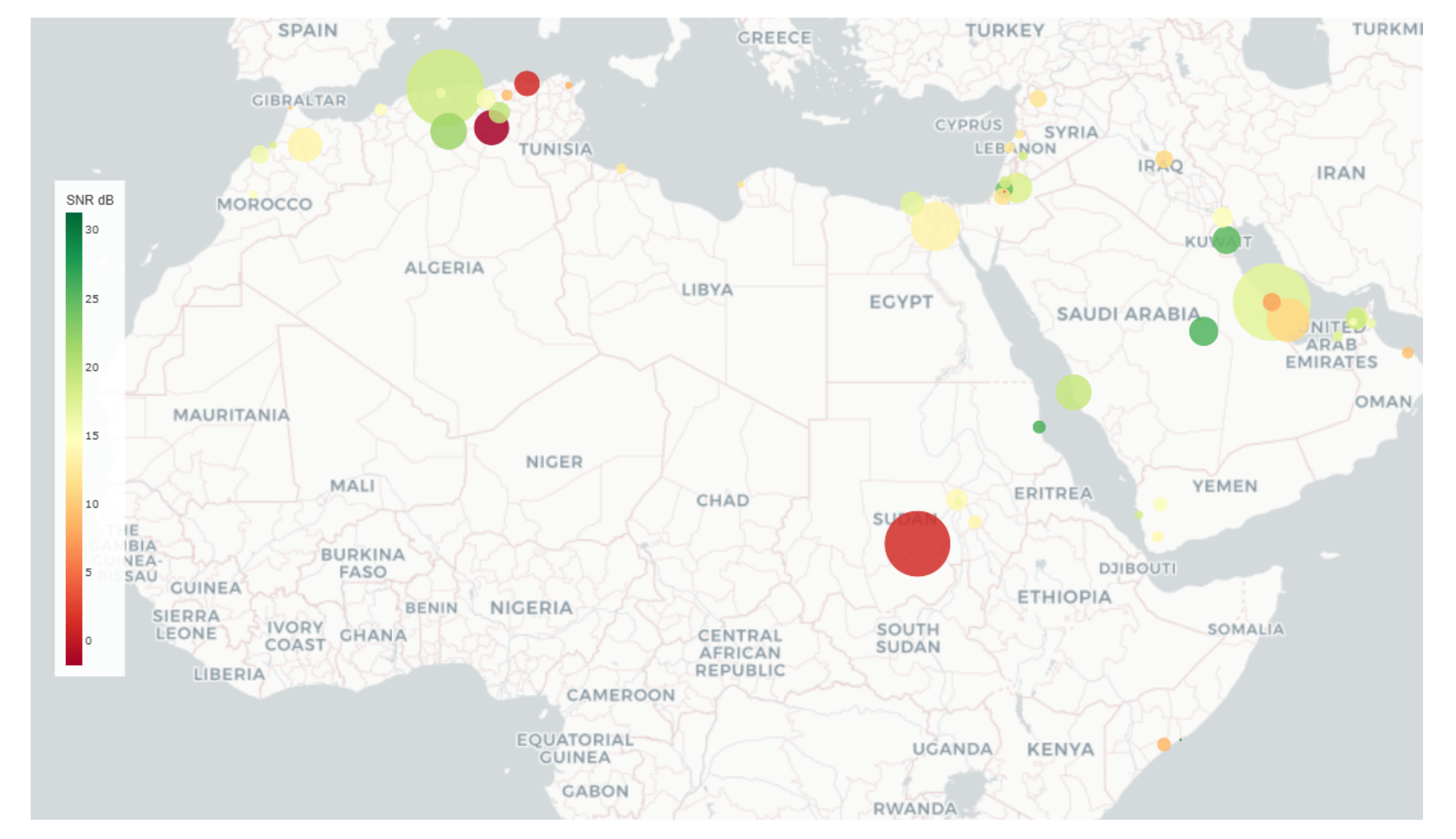}
    \caption{Geographic distribution of the 2{,}329 audio samples across 19 countries and 46 cities. Each circle marks a city: its \emph{size} is proportional to the number of samples collected from that location, while its \emph{color} encodes the mean signal-to-noise ratio (SNR, in dB) of the recordings at that city, as shown by the color bar (red $\approx 0$\,dB indicates noisier audio, green $\approx 30$\,dB cleaner audio). The colour thus summarizes recording quality per location and is independent of the sample-count encoding given by circle size.}
    \label{fig:dataset_map}
\end{figure}

The dataset exhibits substantial geographic diversity with significant sampling-density variations. Algeria contributes the largest share of the validation pool (507 samples), followed by Sudan (290), Egypt (250), Morocco (229), and the United Arab Emirates (212). At the city level, El~Obeid (171), Cairo (152), Algiers (121), Amman (103), and Fes (101) are the densest. Several countries (Qatar, Bahrain, Oman) are represented by fewer than 50 samples, and several cities by only a handful, reflecting both population demographics and practical data-collection constraints. To avoid overfitting to over-represented regions, this imbalance necessitates a careful evaluation methodology, such as our leakage-free GroupKFold split by source recording (Section~\ref{sec:cv-protocol}). Figure~\ref{fig:dataset_map} maps this distribution, with each city sized by sample count and coloured by mean recording SNR, making both the geographic spread and the per-location quality variation visible at a glance.

\subsection{Annotation and Preprocessing}

Because the source material is radio broadcasts, we do not have access to per-speaker self-reported origins. Instead, the ground-truth location of each clip is taken to be the city of its broadcasting station, as recorded in the ARCADE metadata~\cite{nacar2026arcade}; we performed a manual review of the station-to-city assignments and discarded clips with inconsistent or ambiguous geographic metadata. Each (city,~country) pair is then mapped to a coordinate by geocoding against the Nominatim service~\cite{nominatim}, yielding the city-centroid latitude--longitude used as the regression target. This station-city assumption is a deliberate simplification: it treats the dialect of a station's broadcast as representative of its city, which is reasonable for regional stations but introduces label noise for national broadcasters whose presenters may originate elsewhere. All audio is resampled to 16\,kHz mono. We further enforce that no source recording appears in both the training and validation partitions of any fold (Section~\ref{sec:cv-protocol}), so that evaluation reflects geographic generalization rather than recording-level memorization.

\subsection{Data Statistics}

Table~\ref{tab:dataset_stats} summarizes the key properties of the filtered dataset and the parameters of our data-quality pipeline. It records the post-filtering corpus size (2{,}329 samples spanning 19 countries and 46 cities), the most heavily sampled country (Algeria, 507 samples) and city (El~Obeid, 171 samples), which together quantify the geographic imbalance discussed above. The remaining rows document the two thresholds that define our filtering and per-fold label handling --- the rare-city threshold ($<5$ training samples in a fold are merged into the \texttt{\_\_OTHER\_\_} bucket) and the outlier threshold ($>2000$\,km from the country centroid) --- as well as the geocoding source (Nominatim) used to obtain coordinates. These last entries make the dataset reproducible from the raw ARCADE collection.

\begin{table}[h]
\centering
\caption{Key dataset statistics after data-quality filtering.}
\label{tab:dataset_stats}
\begin{tabular}{l c}
\toprule
\textbf{Statistic} & \textbf{Value} \\
\midrule
Total samples (post-DQ) & 2{,}329 \\
Countries & 19 \\
Cities & 46 \\
Largest country (Algeria) & 507 samples \\
Largest city (El~Obeid) & 171 samples \\
Rare-city threshold (per fold) & $<5$ training samples $\to$ \texttt{\_\_OTHER\_\_} \\
Geocoding source & Nominatim (city,~country) lookup \\
DQ outlier threshold & $>2000$\,km from country centroid \\
\bottomrule
\end{tabular}
\end{table}

The dataset represents one of the largest collections of geographically annotated Arabic speech samples, enabling robust evaluation of continuous geolocation approaches.

\section{Proposed Methodology}\label{sec5}

We address Arabic dialect geolocation as continuous localization over the
Earth's surface rather than classification into predefined geographic cells.
Discretization imposes artificial decision boundaries, is sensitive to class
imbalance, and cannot predict at locations unseen during training. These limitations are especially acute for Arabic, whose dialect boundaries are fluid,
overlapping, and shaped by migration, trade routes, and sociolinguistic contact
rather than by administrative borders. The extent to which such variation is in
fact spatially continuous is treated here as a hypothesis to be tested rather
than assumed, and is examined directly in Section~\ref{sec:mantel}.

A continuous formulation alone, however, is not sufficient. Regressing a single
coordinate implicitly assumes that each utterance corresponds to one point,
whereas a dialectal cue is typically compatible with an extended region and, for
shared varieties such as Gulf Arabic, with several mutually distant candidates.
Under such ambiguity a point estimator minimizes its expected error by predicting
the centroid of the plausible set---a location that may itself be implausible. We
therefore model the target as a probability density over the unit sphere and
decode its dominant mode, retaining a continuous output space while allowing the
model to express multimodal geographic uncertainty instead of averaging it away.

Our approach integrates three components: (1) a multi-level acoustic
representation that captures complementary dialectal cues; (2) a two-stage
architecture that decouples frozen pretrained encoders from a trained fusion
backbone; and (3) a multi-task objective coupling a von Mises--Fisher mixture
density with a contrastive location-retrieval head and auxiliary coarse-to-fine
classification, so that the shared representation is organized geographically at
several scales.

\subsection{Model Architecture}
\label{sec:architecture}

\begin{figure*}[!t]
    \centering
    \includegraphics[width=\textwidth]{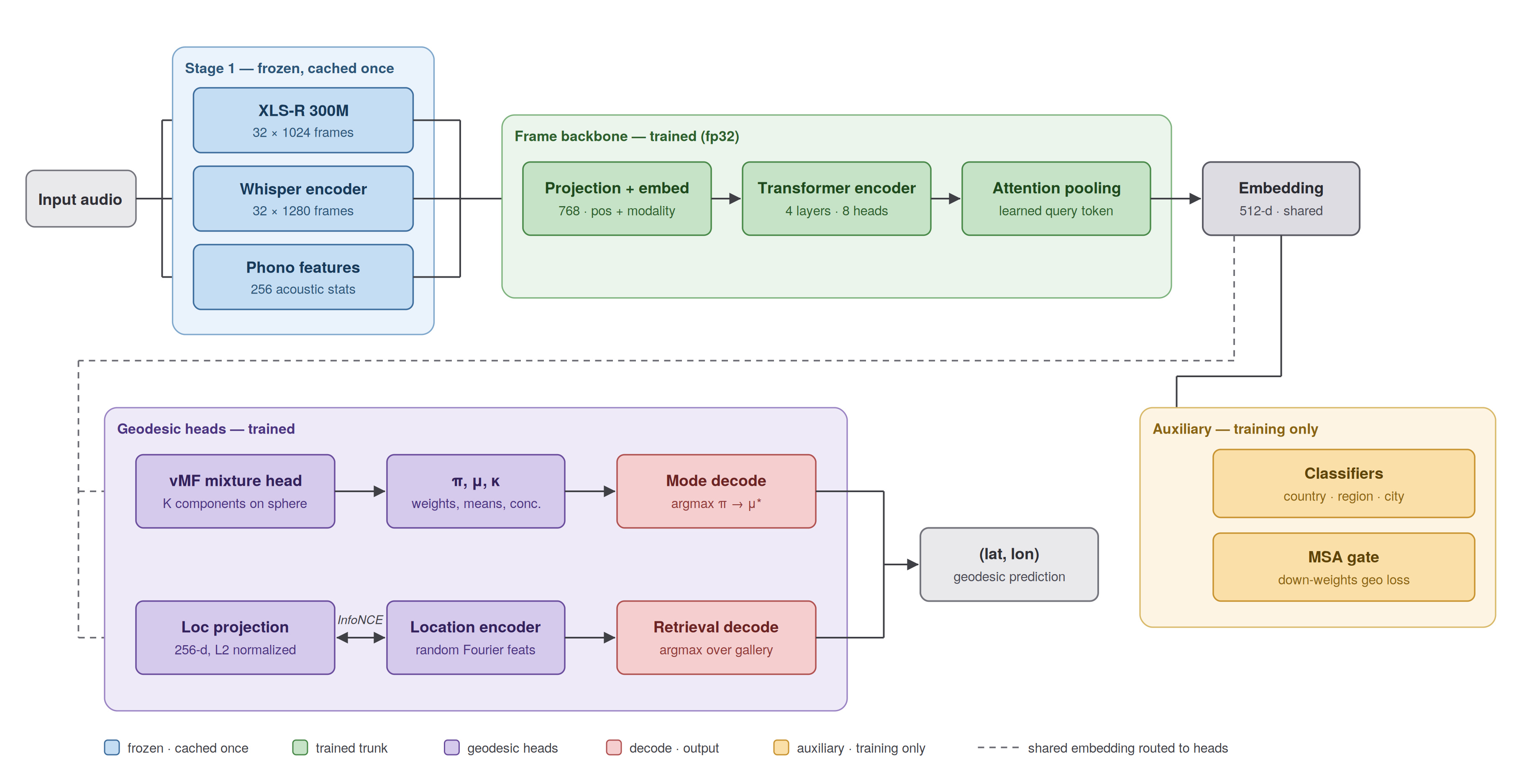}
    \caption{Overview of the proposed Arabic dialect geolocation architecture (GeoArc-FF).
Complementary frame-level features are extracted once from raw speech by three
frozen encoders---XLS-R~\cite{babu2022xlsr}, the Whisper
encoder~\cite{radford2023whisper}, and handcrafted phonological
descriptors---and cached, decoupling the pretrained front-end from optimisation.
A trained backbone projects the three streams into a unified latent space,
augments them with positional and modality embeddings, and fuses them through a
Transformer encoder followed by query-token attention pooling.
The resulting shared 512-d representation is jointly optimised by two geodesic
heads---a von Mises--Fisher mixture density over the unit sphere, decoded by its
dominant mode, and a contrastive projection aligned to a random-Fourier location
encoder~\cite{vivanco2023geoclip} and decoded by retrieval---together with
auxiliary country, region, and city classification heads and an MSA gate that
down-weights the geographic loss on dialect-neutral speech.}
    \label{fig:architecture}
\end{figure*}

Figure~\ref{fig:architecture} presents a high-level overview of the proposed
architecture and its main computational blocks. The design follows a two-stage
pipeline. In the first stage, three complementary frozen encoders are evaluated
once over the corpus and their frame-level outputs cached, so that the
pretrained front-end supplies fixed representations rather than being optimised
jointly. This decoupling is deliberate: with only 2{,}329 dialectal clips,
fine-tuning a 300M-parameter encoder is prone to overfitting, and caching renders
the five-fold masked-city cross-validation protocol computationally tractable
while isolating the contribution of the head design. In the second stage, a
trained backbone projects the three streams into a unified latent space, fuses
them with a Transformer encoder over the concatenated token sequence, and pools
them into a single utterance-level representation. This shared representation is
read by parallel prediction heads: two geodesic heads that jointly produce the
coordinate, and auxiliary classification heads that shape the latent space during
training without contributing to inference. The remainder of this section details
each component in isolation, emphasising its role in capturing complementary
dialectal cues and in modelling the regional---rather than point-like---nature of
dialect geography.

\subsubsection{Audio Preprocessing and Feature Extraction}

All audio signals are resampled to 16\,kHz mono and centred to a fixed 10-second window (zero-padded if shorter, centre-cropped if longer) before peak normalization. Dialectal variation manifests across multiple linguistic levels, ranging from fine-grained phonetic realizations to broader prosodic and rhythmic patterns. No single representation captures all these phenomena effectively. We therefore adopt a multi-representation strategy that explicitly targets different aspects of dialectal variation, retaining \emph{frame-level} (rather than utterance-pooled) features for two of the three modalities so that downstream attention can attend to local cues.

\paragraph{Self-Supervised Speech Representations.}
We extract contextual frame embeddings using XLS-R-300M (\texttt{facebook/wav2vec2-xls-r-300m}). Self-supervised models learn rich phonetic and lexical representations from large volumes of unlabeled multilingual speech, making them particularly suitable for dialectal settings. The native temporal sequence is downsampled by adaptive average pooling to a fixed length of $T = 32$ frames, yielding a tensor in $\mathbb{R}^{32 \times 1024}$ per utterance.

\paragraph{Whisper Encoder Features.}
To complement self-supervised features we incorporate the encoder of Whisper-large-v3 (\texttt{openai/whisper-large-v3}). Whisper is trained on large-scale multilingual and multi-domain speech data, enabling it to capture language-agnostic acoustic patterns and robust temporal structures. The decoder is discarded to save memory; encoder hidden states are again downsampled to $T = 32$ frames, yielding $\mathbb{R}^{32 \times 1280}$.

\paragraph{Phonotactic and Prosodic Features.}
While neural embeddings capture high-level abstractions, hand-crafted acoustic features provide complementary cues related to speech rhythm, timbre, and articulation. We compute MFCCs (20 coefficients with first and second derivatives, summarized by per-coefficient mean/std/max/min), spectral descriptors (centroid, bandwidth), zero-crossing rate, and a 64-bin mel spectrogram time-average. The concatenation, padded/truncated to $\mathbb{R}^{1000}$, forms a single utterance-level vector $\mathbf{p}$.

\subsubsection{Feature Projection and Fold-Local Normalization}

The phonotactic vector $\mathbf{p}$ is stored \emph{raw} in our feature cache. At training time, fold-local normalization statistics ($\mu, \sigma$) are computed on the training indices of the current cross-validation fold only, applied at the data-loader level, and finally rescaled by a learnable \texttt{LayerNorm} inside the backbone. This explicit fold-local protocol replaces the global $z$-score that we identified as a source of evaluation leakage in earlier iterations of this work. Each modality is then projected to a shared $d = 768$ hidden dimension via a per-modality \texttt{LayerNorm} followed by a linear layer.

\subsubsection{Frame-Level Backbone with Attention Pool}

The 32 XLS-R frame tokens, 32 Whisper-encoder frame tokens, and the single phonotactic token are concatenated along the time axis into a sequence of length $2T + 1 = 65$ tokens. Learned positional embeddings and a learned three-way modality embedding are added to disambiguate the source of each token. The sequence is processed by a 4-layer Transformer encoder (8 heads, $d_{\text{ff}} = 4d$, GELU, pre-norm) which models long-range dependencies across modalities and time.

\paragraph{Learnable-Query Attention Pool.}
A single learnable query token $\mathbf{q} \in \mathbb{R}^{1 \times d}$ attends over the entire encoded sequence through a multi-head attention block, yielding the utterance representation $\mathbf{h}_{\text{pool}} \in \mathbb{R}^d$. This attention-pool mechanism plays the role of a soft \texttt{[CLS]} token: it lets the network adaptively weight phonetic, prosodic, and self-supervised cues per utterance rather than relying on a fixed mean-pool.

\subsubsection{Multi-Task Prediction Heads}

Geographic variation exhibits a natural hierarchy, progressing from broad regions to fine-grained local variation. To exploit this structure, the model jointly predicts continuous geographic coordinates alongside auxiliary country and city classifications.

\paragraph{Geographic Regression Head.}
Geographic location is predicted as a unit-norm 3D Cartesian vector:
\begin{equation}
\mathbf{p}_{\text{geo}} = \frac{\mathbf{W}_{\text{reg}}\,\mathbf{h}_{\text{pool}}}{\left\| \mathbf{W}_{\text{reg}}\,\mathbf{h}_{\text{pool}} \right\|}.
\end{equation}
Unit normalization constrains predictions to Earth's surface and simplifies distance computation. Country and city heads are simple two-layer MLPs over $\mathbf{h}_{\text{pool}}$.

\paragraph{Country and City Classification Heads.}
Auxiliary country and city classifiers provide hierarchical supervision:
\begin{equation}
\mathbf{p}_{\text{country}} = \mathrm{softmax}(\mathbf{W}_{\text{country}} \mathbf{h}_{\text{pool}}),
\end{equation}
\begin{equation}
\mathbf{p}_{\text{city}} = \mathrm{softmax}(\mathbf{W}_{\text{city}} \mathbf{h}_{\text{pool}}).
\end{equation}
These auxiliary tasks regularize the shared representation and encourage geographic consistency.

\subsection{Loss Functions and Training Objectives}

\subsubsection{Spherical Geodesic Loss}

We optimize angular distance on the unit sphere:
\begin{equation}
\mathcal{L}_{\text{angular}} = \frac{1}{N} \sum_{i=1}^{N} \arccos \left( \mathrm{clamp}(\mathbf{p}_i \cdot \mathbf{y}_i, -0.99999, 0.99999) \right).
\end{equation}
Angular distance corresponds directly to great-circle distance, aligning the training objective with real-world geographic error. The clamping operation ensures numerical stability near $\pm 1$.

\subsubsection{Multi-Objective Loss Function}

The total loss combines regression and classification objectives:
\begin{equation}
\mathcal{L}_{\text{total}} = \alpha \mathcal{L}_{\text{angular}} + \beta \mathcal{L}_{\text{country}} + \gamma \mathcal{L}_{\text{city}},
\end{equation}
with $\alpha = 0.5$, $\beta = 0.5$, and $\gamma = 0.4$. These weights balance fine-grained localization against the two hierarchical classification objectives. Cross-entropy with label smoothing of $0.05$ is used for both classifiers, and MixUp~/~SLERP-augmented batches use the corresponding soft targets.

\subsection{Training Strategy}

\subsubsection{Augmentation, EMA and Snapshot Ensembling}

During training we apply (i)~SpecAugment-style time and channel masking on the XLS-R and Whisper frame sequences, (ii)~MixUp ($\alpha = 0.2$) on inputs and on classification targets, with SLERP interpolation on the unit-sphere coordinates so that mixed targets remain valid geographic points, and (iii)~Gaussian latitude/longitude jitter ($\sigma = 0.05^\circ$) on the regression targets. Model weights are tracked by an exponential moving average (decay $0.999$); each evaluation uses the EMA weights. We further apply test-time augmentation by averaging predictions across five passes with input Gaussian noise ($\sigma = 0.02$).

A snapshot ensemble retains the top-3 EMA snapshots over training, subject to a minimum gap of 10 epochs between retained snapshots; final per-fold predictions average the unit-norm coordinates and the country/city softmaxes from these snapshots.

\subsubsection{Optimization and Implementation Details}

Training employs the configuration shown in Table~\ref{tab:training_hyperparams}. Models are trained on a single NVIDIA P100 / T4 / A100 GPU (Kaggle pool); a fold typically converges within 40--80 epochs.
\begin{table}[h]
\centering
\caption{Training hyperparameters and optimization settings.}
\label{tab:training_hyperparams}
\begin{tabular}{l p{8cm}}
\toprule
\textbf{Hyperparameter} & \textbf{Setting} \\
\midrule
Optimizer & AdamW ($\beta_1=0.9$, $\beta_2=0.999$, weight decay $5{\times}10^{-5}$) \\
Learning rate schedule & Initial LR $10^{-4}$ with cosine annealing to $10^{-6}$ \\
Batch size & 24 \\
Gradient clipping & Maximum norm of 1.0 \\
Max epochs / patience & 80 / 18 \\
EMA decay & 0.999 \\
MixUp $\alpha$ & 0.2 (SLERP on $xyz$) \\
TTA passes / noise $\sigma$ & 5 / 0.02 \\
Snapshot ensemble & top-3, min gap 10 epochs \\
\bottomrule
\end{tabular}
\end{table}

\subsubsection{Leakage-Free Cross-Validation Protocol}
\label{sec:cv-protocol}

All evaluation is performed under a 5-fold \texttt{GroupKFold} split where the grouping key is the source recording identifier (extracted from the audio filename). Because the ARCADE corpus is built from segmented radio recordings, multiple short clips can share a parent recording; grouping by source identifier prevents the same recording from contributing to both training and validation. We additionally enforce that all preprocessing statistics (the phonotactic mean/variance and both \texttt{LabelEncoder} vocabularies for country and city) are fit on the training indices of the current fold only. An explicit \texttt{\_\_OTHER\_\_} bucket at index $0$ catches validation labels that fall outside the per-fold training vocabulary, including cities pruned by the per-fold rare-city threshold ($<5$ training samples).

\paragraph{City-masking for unseen-city evaluation.}
Fold~$0$ serves as a baseline reference (all cities visible during training, modulo the rare-city threshold). For folds $1$ through $4$, we additionally remove all training samples of two randomly chosen cities; those cities still appear in that fold's validation set, providing a true zero-shot, unseen-city evaluation. The choice of masked cities is deterministic (seed $20260427 + \text{fold\_id}$) and is constrained so that (i)~the masked cities appear in both the train and val partitions of the original split, and (ii)~the post-mask training set retains at least eight unique cities. The resulting per-fold masking is recorded in \texttt{cv\_plan.json} and reproduced in Table~\ref{tab:per_fold} (column \textsc{masked}).

\section{Experimental Evaluation}\label{sec6}

\subsection{Evaluation Metrics}

\subsubsection{Geodesic Distance Error}

The primary evaluation metric is the geodesic distance between predicted and ground-truth locations measured along the Earth's surface. Predictions are represented as unit vectors on the sphere to avoid distortions inherent to planar latitude-longitude regression. Given predicted and ground-truth vectors $\mathbf{p}$ and $\mathbf{y}$, the angular distance is computed as:
\begin{equation}
d_{\text{ang}} = \arccos(\mathbf{p} \cdot \mathbf{y}),
\end{equation}
and converted to kilometers by:
\begin{equation}
d_{\text{geo}} = R \cdot d_{\text{ang}},
\end{equation}
where $R = 6371$ km is the mean Earth radius.

We report the mean, median, and standard deviation of geodesic error. Median error is emphasized due to the heavy-tailed nature of localization errors, where a small number of extreme failures can disproportionately affect the mean.

\subsubsection{Accuracy within Distance Thresholds}

To facilitate interpretability and comparison with prior work, we report accuracy within predefined distance thresholds:
\begin{equation}
\text{Acc@}r = \mathbb{P}(d_{\text{geo}} \le r),
\end{equation}
with $r \in \{50, 100, 250, 500, 1000, 2500\}$ km. These thresholds correspond to intra-city, inter-city, regional, national, multi-country, and continental localization scales, respectively.

\subsubsection{Country and City Classification Accuracy}

For the auxiliary hierarchical heads, we report top-1 accuracy for both country and city prediction. We additionally summarize per-fold accuracy as mean$\pm$standard deviation and the corresponding 95\% bootstrap confidence interval over the held-out folds.

\subsubsection{Calibration and Confidence Quality}
We assess probabilistic calibration of the auxiliary classifiers using Expected
Calibration Error (ECE) and the Brier score. Let $n$ denote the number of
evaluation clips, $K$ the number of classes, $\hat{p}_{ik}$ the predicted
probability that clip $i$ belongs to class $k$, and $y_i$ the ground-truth
label. We write $\hat{y}_i = \arg\max_k \hat{p}_{ik}$ for the predicted label
and $\hat{c}_i = \max_k \hat{p}_{ik}$ for its associated confidence.

ECE measures the difference between confidence and accuracy across binned
predictions. Partitioning the clips into $M$ equal-width confidence bins
$B_1,\dots,B_M$ over $[0,1]$ according to $\hat{c}_i$, ECE is the
sample-weighted mean absolute gap between per-bin accuracy and confidence:
\begin{equation}
  \mathrm{ECE} = \sum_{m=1}^{M} \frac{|B_m|}{n}
    \bigl\lvert \mathrm{acc}(B_m) - \mathrm{conf}(B_m) \bigr\rvert ,
\end{equation}
where
\begin{equation}
  \mathrm{acc}(B_m) = \frac{1}{|B_m|} \sum_{i \in B_m} \mathbf{1}[\hat{y}_i = y_i] ,
  \qquad
  \mathrm{conf}(B_m) = \frac{1}{|B_m|} \sum_{i \in B_m} \hat{c}_i .
\end{equation}
We use $M = 15$ bins and omit empty bins from the sum. A perfectly calibrated
classifier satisfies $\mathrm{acc}(B_m) = \mathrm{conf}(B_m)$ for all $m$,
yielding $\mathrm{ECE} = 0$.

The Brier score quantifies the squared difference between the predicted
probability vector and the one-hot encoding of the true outcome:
\begin{equation}
  \mathrm{BS} = \frac{1}{n} \sum_{i=1}^{n} \sum_{k=1}^{K}
    \bigl( \hat{p}_{ik} - \mathbf{1}[y_i = k] \bigr)^2 .
\end{equation}
Whereas ECE inspects only the top-label confidence, the Brier score is a
strictly proper scoring rule that penalises the entire predictive
distribution, jointly rewarding calibration and sharpness. Under this
unnormalised, sum-over-classes convention $\mathrm{BS} \in [0, 2]$, with $0$
attained when unit probability mass is placed on the correct class. Both
quantities are computed per fold and averaged across the five folds.

Calibration quality is critical for uncertainty-aware downstream applications
such as selective prediction, human-in-the-loop verification, and
risk-sensitive deployment scenarios.

\subsubsection{Selective Prediction.}

We report selective-prediction operating points by sweeping the per-sample maximum-softmax confidence (separately for the country and the city head) and reporting the threshold and pooled metrics that achieve $80\%$ coverage of the validation set.

\subsubsection{Mantel Test on the Latent Space.}

To quantify the relationship between linguistic and geographic distance, we run a permutation Mantel test on $1{,}124{,}250$ validation pairs (10{,}000 permutations), correlating Euclidean distance in $\mathbf{h}_{\text{pool}}$ with great-circle distance between ground-truth coordinates.

\subsubsection{Border vs. Interior Decomposition.}\label{sec:border}

We additionally split the validation set into border and interior samples. In the absence of a country-polygon library on the training environment, we use a centroid-based heuristic: a sample is labelled \emph{border} if any \emph{other} country's centroid is closer than $150$\,km to the sample's true coordinates. We report mean error, median error, and country accuracy for each subset, and treat the resulting numbers as an indicative rather than a definitive boundary analysis.

\subsubsection{Unseen-City Zero-Shot Evaluation.}

The aim of this evaluation is to answer the single question that standard cross-validation cannot: \emph{how well does the model localize speech from a city it has never seen during training?} Under an ordinary shuffled split, every city in the validation set is also present in the training set, so the model can succeed by memorizing city-specific acoustic signatures rather than learning the underlying geographic--dialectal structure. The resulting metrics therefore conflate genuine geographic generalization with city-level recall and systematically overstate deployable performance, since any real-world system will inevitably encounter speakers from locations absent in its training data. To isolate true generalization, for folds 1--4 we mask two cities per fold from the training set while retaining their samples in validation (Section~\ref{sec:cv-protocol}), and evaluate these held-out cities separately from the seen cities of the same fold. We report the seen/unseen mean-error ratio as a direct measure of zero-shot geographic generalization: a ratio near $1$ indicates that the model interpolates smoothly to unseen locations from neighbouring dialect regions --- exactly the behaviour predicted by the continuum hypothesis --- whereas a large ratio reveals reliance on having seen the target city. This protocol provides a leakage-free lower bound on real-world performance and, as Section~\ref{sec:unseen-city} shows, exposes how strongly zero-shot accuracy depends on how densely the surrounding dialect region is sampled.

\section{Results and Discussion}\label{sec7}

All numbers reported in this section are pooled across the five validation folds described in Section~\ref{sec:cv-protocol}; per-fold means and 95\% bootstrap confidence intervals are given alongside the pooled value where appropriate.

\subsection{Overall Geolocation Performance}

\begin{figure}[H]
\centering
\begin{minipage}[t]{0.56\textwidth}
    \centering
    \vspace{0pt}
    \includegraphics[width=\textwidth]{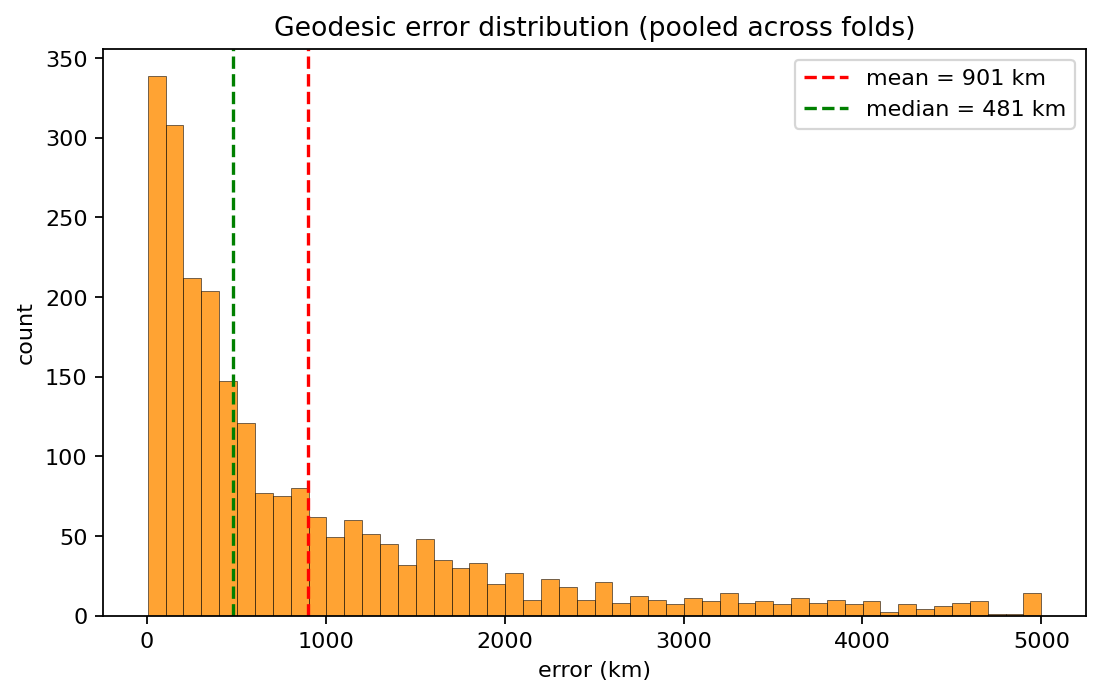}
\end{minipage}
\hfill
\begin{minipage}[t]{0.43\textwidth}
    \centering
    \vspace{30pt}
    \footnotesize
    \setlength{\tabcolsep}{3pt}
    \begin{tabular}{lcc}
        \toprule
        \textbf{Metric} & \textbf{Value} & \textbf{Unit} \\
        \midrule
        Mean Error          & 901.5   & km \\
        Median Error        & 481.2   & km \\
        25\textsuperscript{th} percentile & 179.9  & km \\
        75\textsuperscript{th} percentile & 1228.2 & km \\
        90\textsuperscript{th} percentile & 2340.5 & km \\
        Mean (5-fold, 95\% CI) & $[854.6, 969.1]$ & km \\
        Median (5-fold, 95\% CI) & $[445.3, 497.0]$ & km \\
        \bottomrule
    \end{tabular}
\end{minipage}
\caption{Distribution of geodesic localization errors on the pooled validation set and corresponding summary statistics. The error distribution is strongly right-skewed: the median (481\,km) sits well below the mean (902\,km) and the 90\textsuperscript{th} percentile (2341\,km), reflecting a long tail of high-error samples.}
\label{fig:error_distribution_and_metrics}
\end{figure}

The proposed model attains a pooled mean localization error of $901.5$\,km and a pooled median of $481.2$\,km on the held-out validation samples (one prediction per source recording, aggregated across folds). The wide gap between mean and median, together with the $90\%$-percentile error of $2340.5$\,km, confirms the heavy-tailed nature of geographic prediction: the network localizes a clear majority of utterances to within a few hundred kilometres, but a long tail of confusable samples (low-content recordings, code-switching, or dialects with weak distinctive cues) drives the mean upward. The fold-level mean is $901.5 \pm 77.0$\,km (95\% CI $[854.6, 969.1]$) and the fold-level median is $472.8 \pm 32.9$\,km (95\% CI $[445.3, 497.0]$), indicating that fold-to-fold variability is small and the pooled numbers are stable. Figure~\ref{fig:error_distribution_and_metrics} plots the full distribution of geodesic errors alongside these summary statistics, and Table~\ref{tab:geolocation_performance} lists the headline metrics.

\begin{figure}[H]
\centering
\begin{minipage}[t]{0.52\textwidth}
    \centering
    \vspace{0pt}
    \includegraphics[width=\textwidth]{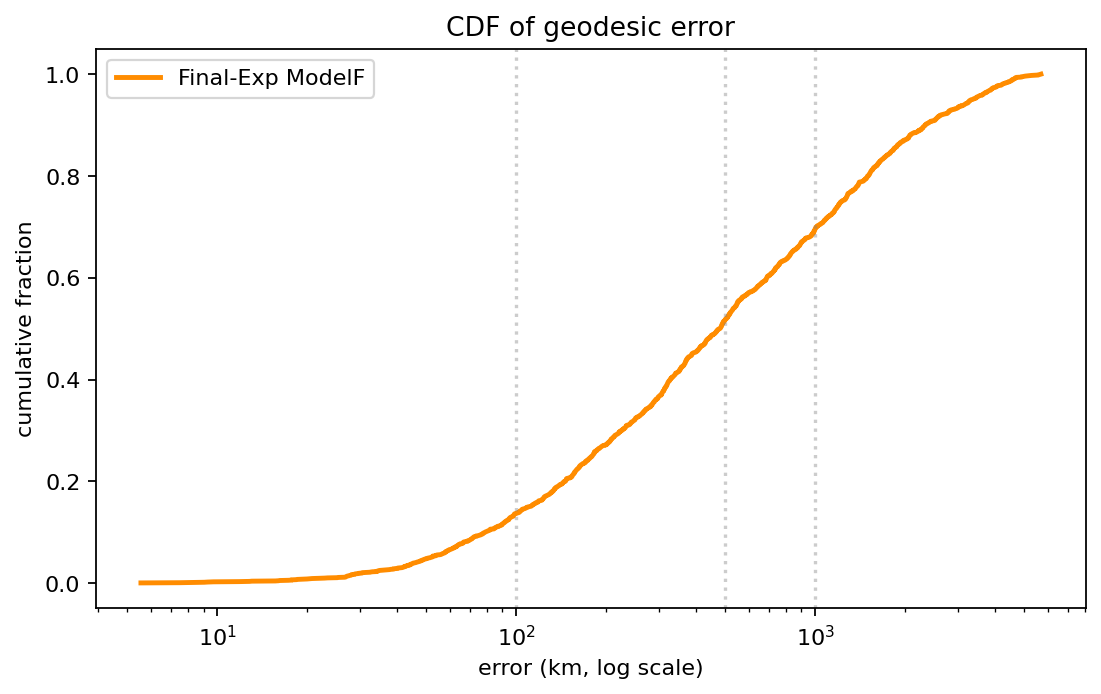}
\end{minipage}
\hfill
\begin{minipage}[t]{0.47\textwidth}
    \centering
    \vspace{35pt}
    \small
    \begin{tabular}{lcc}
        \toprule
        \textbf{Threshold} & \textbf{Accuracy} & \textbf{Incremental Gain} \\
        \midrule
        $< 50$\,km   &  4.8\% &  4.8\% \\
        $< 100$\,km  & 13.7\% &  8.9\% \\
        $< 250$\,km  & 32.2\% & 18.5\% \\
        $< 500$\,km  & 51.7\% & 19.5\% \\
        $< 1000$\,km & 69.5\% & 17.8\% \\
        $< 2500$\,km & 90.9\% & 21.4\% \\
        \bottomrule
    \end{tabular}
\end{minipage}
\caption{Cumulative distribution of localization errors and accuracy within increasing distance thresholds.}
\label{fig:cdf_and_threshold_accuracy}
\end{figure}

Threshold accuracies confirm that performance scales smoothly across geographic resolutions. As Figure~\ref{fig:cdf_and_threshold_accuracy} shows, $13.7\%$ of predictions fall within $100$\,km, $51.7\%$ within $500$\,km, $69.5\%$ within $1000$\,km, and $90.9\%$ within $2500$\,km. These numbers establish that the model reliably places most utterances within the correct multi-country region, while sub-city precision remains difficult under leakage-free evaluation. The auxiliary heads attain pooled top-1 accuracies of $64.5\%$ for country (5-fold mean $0.645 \pm 0.044$, 95\% CI $[0.608, 0.677]$) and $45.2\%$ for city ($0.452 \pm 0.050$, CI $[0.415, 0.489]$).

\begin{table}[H]
\centering
\caption{Headline geolocation performance, pooled across 5-fold GroupKFold validation.}
\label{tab:geolocation_performance}
\begin{tabular}{lcc}
\toprule
\textbf{Metric} & \textbf{Value} & \textbf{Unit} \\
\midrule
Pooled mean error           & 901.5 & km \\
Pooled median error         & 481.2 & km \\
Country accuracy (pooled)   & 64.5  & \% \\
City accuracy (pooled)      & 45.2  & \% \\
\bottomrule
\end{tabular}
\end{table}

\subsubsection{Per-Fold Decomposition}

\begin{figure}[H]
    \centering
    \includegraphics[width=0.95\textwidth]{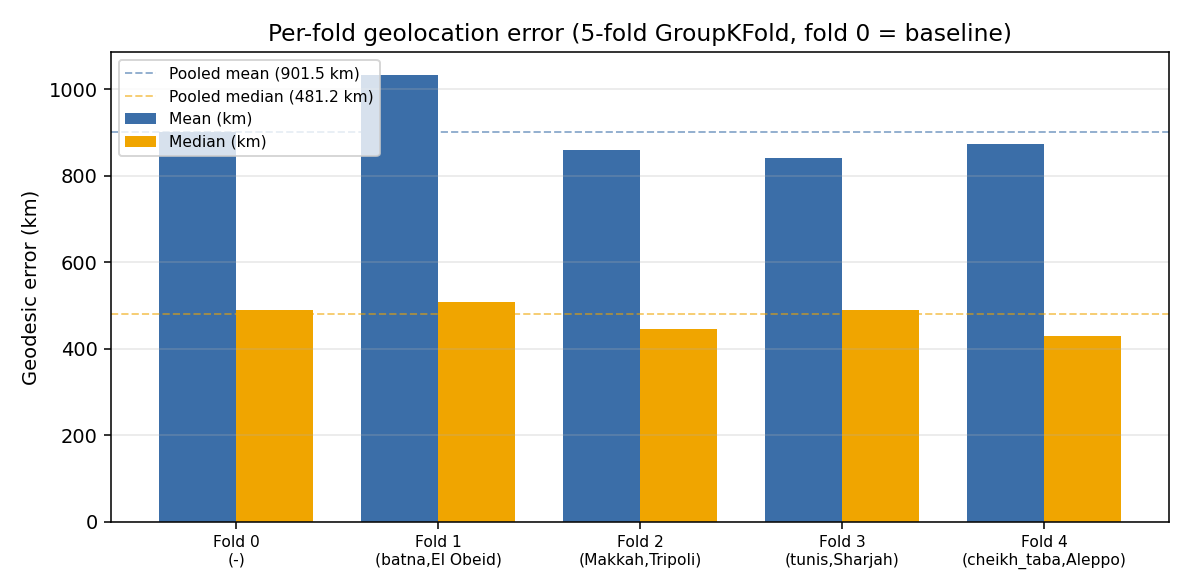}
    \caption{Per-fold mean and median geodesic error. Dashed lines indicate the pooled mean ($901.5$\,km) and median ($481.2$\,km). The remarkable stability of the median across folds ($\pm 33$\,km) contrasts with the heavier-tailed mean ($\pm 77$\,km) and reflects the heavy-tailed nature of geodesic error.}
    \label{fig:per_fold_error}
\end{figure}

To make the city-masking protocol reproducible, Table~\ref{tab:per_fold} reports per-fold validation size, masked cities, and pooled mean/median error. The five folds are remarkably consistent in median error (range $430$--$507$\,km, std only $33$\,km), while the mean error is dominated by the heavy tail and varies more (range $840$--$1033$\,km). Folds with masked cities incur slightly larger mean errors than the unmasked baseline (fold~0) on average --- a foretaste of the unseen-city analysis in Section~\ref{sec:unseen-city}. We also include in Figure~\ref{fig:per_fold_error} a side-by-side bar visualization of these per-fold means and medians against the pooled value, which makes the small fold-to-fold spread visible at a glance.

\begin{table}[H]
\centering
\caption{Per-fold validation size, cities masked from training, pooled mean/median geodesic error, and auxiliary head accuracies. The 5-fold means $\pm$ std are $901.5 \pm 77.0$\,km (mean), $472.8 \pm 32.9$\,km (median), $0.645 \pm 0.044$ (country), and $0.452 \pm 0.050$ (city).}
\label{tab:per_fold}

\footnotesize
\setlength{\tabcolsep}{3pt}

\begin{tabular}{@{}c l c c c c c@{}}
\toprule
\textbf{Fold} &
\textbf{Masked cities} &
\textbf{$n_{\mathrm{val}}$} &
\textbf{Mean} &
\textbf{Median} &
\textbf{Country} &
\textbf{City} \\
&
&
&
\textbf{(km)} &
\textbf{(km)} &
\textbf{acc.} &
\textbf{acc.} \\
\midrule
0 & --                  & 466 & 901.1  & 490.3 & 0.691 & 0.474 \\
1 & Batna, El~Obeid     & 466 & 1033.3 & 507.0 & 0.663 & 0.412 \\
2 & Makkah, Tripoli     & 466 & 860.0  & 446.1 & 0.618 & 0.461 \\
3 & Tunis, Sharjah      & 466 & 840.1  & 490.5 & 0.584 & 0.395 \\
4 & Cheikh Taba, Aleppo & 465 & 872.7  & 430.0 & 0.671 & 0.518 \\
\midrule
\textbf{Pooled} &
-- &
\textbf{2329} &
\textbf{901.5} &
\textbf{481.2} &
\textbf{0.645} &
\textbf{0.452} \\
\bottomrule
\end{tabular}
\end{table}

\subsection{Country and City-Level Analysis}

\subsubsection{Country-Level Performance}

\begin{figure}[htbp]
    \centering
    \includegraphics[width=\textwidth]{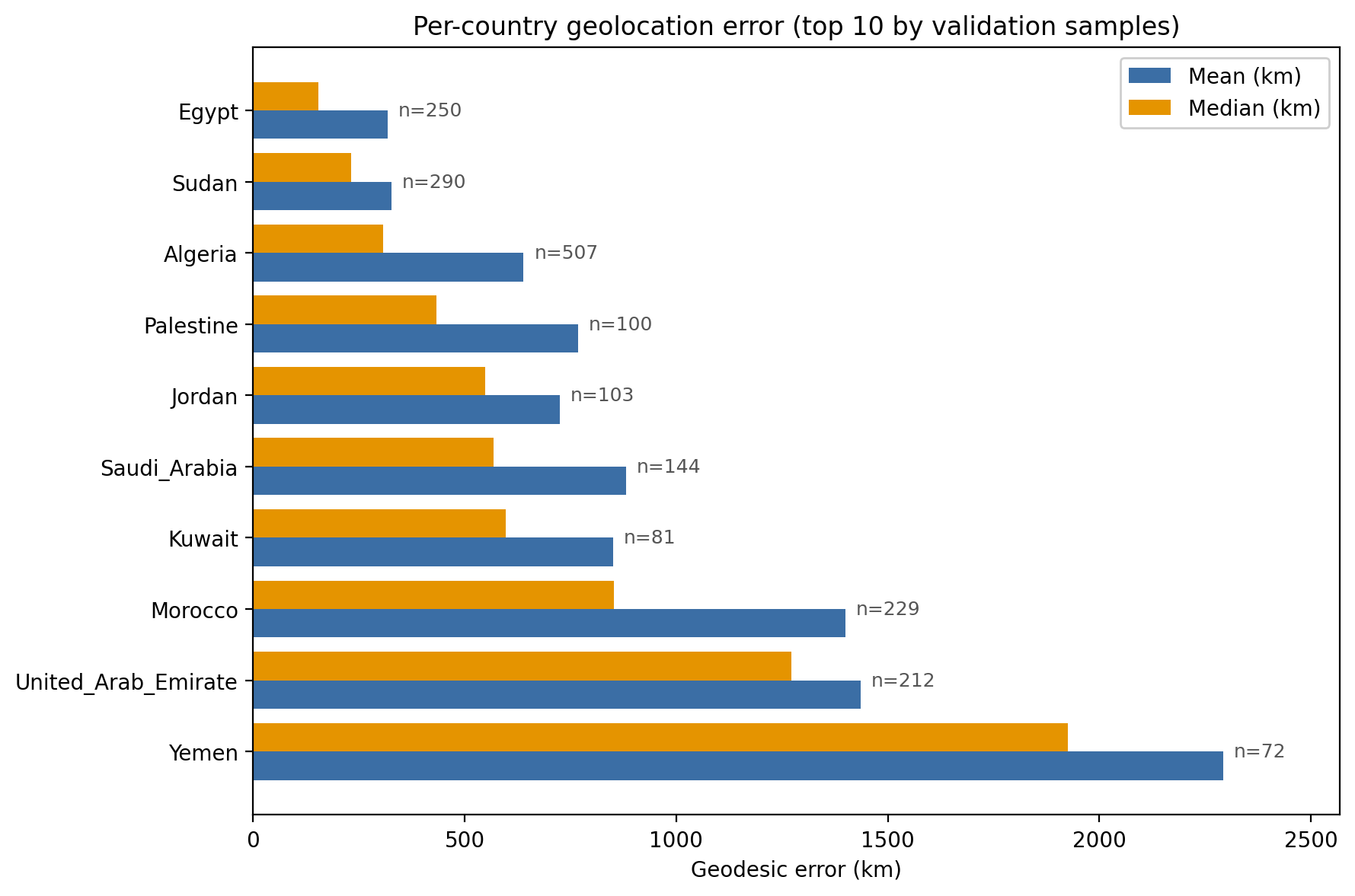}
    \caption{Per-country geodesic error (top 10 by validation samples), sorted by median. Sample-count annotations make the strong dependence on sample density explicit: the densely sampled Egypt/Sudan/Algeria triplet dominates the leaderboard, while sparser Maghrebi and Gulf countries trail by a wide margin.}
    \label{fig:country_error}
\end{figure}

Per-country performance varies by an order of magnitude (Table~\ref{tab:country_performance}). The model performs best on countries with dense, geographically compact city distributions, including Egypt (mean $317.2$\,km, median $153.6$\,km), Sudan (mean $326.9$\,km, median $232.2$\,km) and Algeria (mean $639.2$\,km, median $307.1$\,km despite its $507$-sample size and large within-country dispersion). Maghrebi and Gulf states with broader internal dialectal variation, such as Morocco (mean $1399.3$\,km) and the United Arab Emirates (mean $1435.2$\,km), show clearly weaker localization. Yemen (mean $2293.3$\,km, median $1925.8$\,km) is the worst-performing top-10 country, with most samples mislocated outside its borders entirely. Figure~\ref{fig:country_error} visualizes these per-country mean and median errors as horizontal bars sorted by median, making the broad three-tier picture (Egypt/Sudan; Algeria/Jordan/Palestine/Kuwait/Saudi; Morocco/UAE/Yemen) immediately apparent.

\begin{table}[H]
\centering
\caption{Country-level geolocation performance (top 10 by validation samples). Acc@500 reports the fraction of samples localized within $500$\,km of ground truth.}
\label{tab:country_performance}
\begin{tabular}{lcccc}
\toprule
\textbf{Country} & \textbf{Samples} & \textbf{Mean (km)} & \textbf{Median (km)} & \textbf{Acc@500} \\
\midrule
Algeria       & 507 &  639.2 &  307.1 & 0.69 \\
Sudan         & 290 &  326.9 &  232.2 & 0.82 \\
Egypt         & 250 &  317.2 &  153.6 & 0.83 \\
Morocco       & 229 & 1399.3 &  852.2 & 0.32 \\
UAE           & 212 & 1435.2 & 1272.5 & 0.24 \\
Saudi~Arabia  & 144 &  893.3 &  633.7 & 0.39 \\
Jordan        & 103 &  725.4 &  547.6 & 0.46 \\
Palestine     & 100 &  768.0 &  432.4 & 0.52 \\
Kuwait        &  81 &  850.3 &  596.2 & 0.44 \\
Yemen         &  72 & 2293.3 & 1925.8 & 0.15 \\
\bottomrule
\end{tabular}
\end{table}

\subsubsection{City-Level Performance}

\begin{figure}[H]
    \centering
    \includegraphics[width=\textwidth]{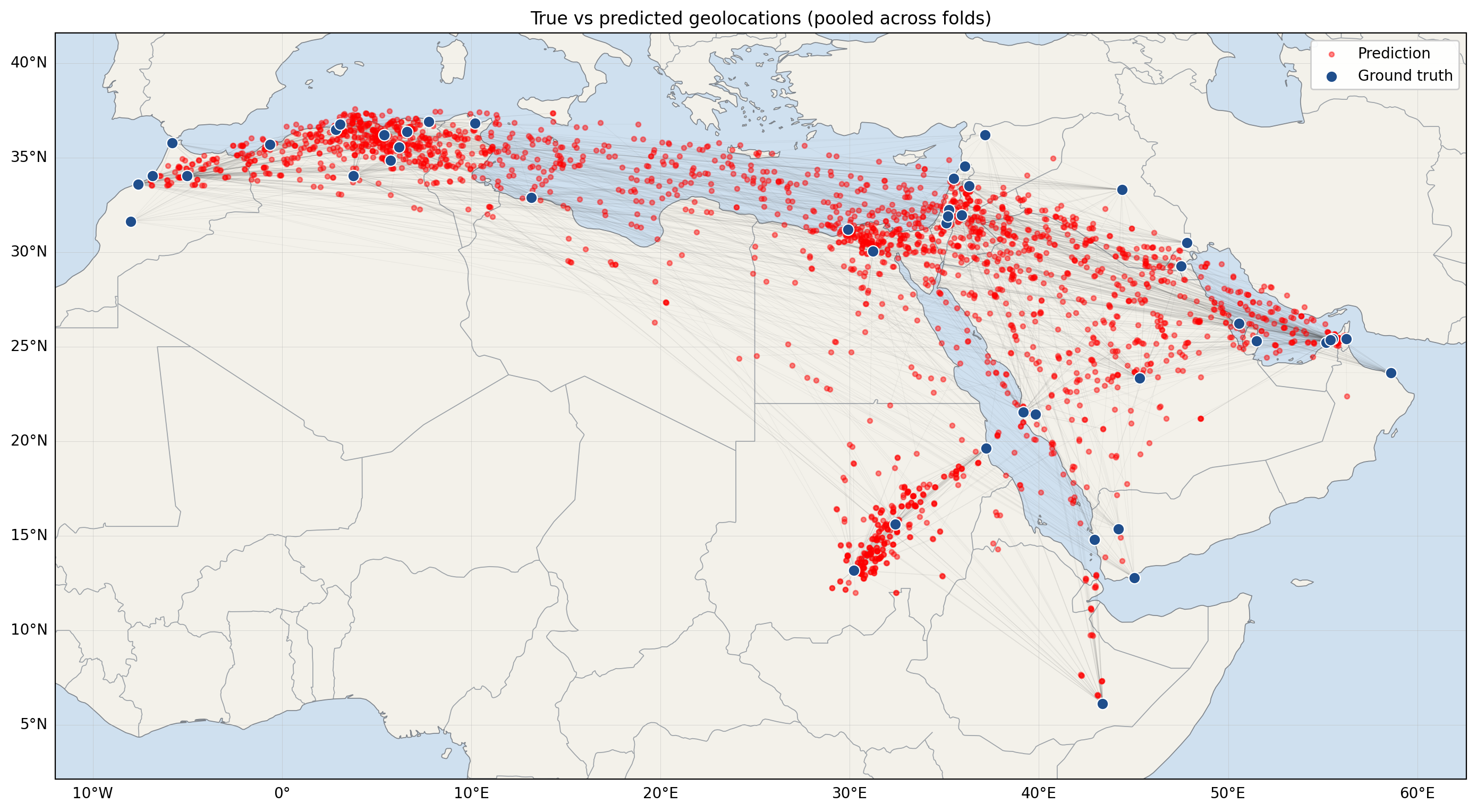}
    \caption{Geographic visualization of city-level audio geolocation performance on the pooled validation set. Blue markers indicate ground-truth city locations and red markers denote model-predicted locations; line segments connect each pair, making per-region error patterns directly visible.}
    \label{fig:city_geolocation_map}
\end{figure}

City-level results echo the country picture (Table~\ref{tab:city_performance}). The strongest cities are Alexandria ($182.5$\,km mean, $97.9$\,km median --- the only city with sub-100\,km median), El~Obeid in Sudan ($291.8$\,km mean, $203.3$\,km median), Omdurman ($332.9$\,km), and Cairo ($404.0$\,km), all densely sampled and dialectally distinctive. Conversely, Fes ($1285.8$\,km), Ajman ($1250.1$\,km) and Sharjah ($1373.7$\,km) sit at the opposite end: Fes is consistently confused with other Moroccan cities, while the two UAE cities are dragged toward each other and toward Saudi alternatives. Algiers' relatively large mean ($792.5$\,km) coexists with a much lower median ($338.4$\,km), an unmistakable signature of a small number of long-distance failure modes inflating the average. Figure~\ref{fig:city_geolocation_map} visualizes these city-level errors on the map, connecting each ground-truth location to its predicted counterpart so that the regional pattern of confusions is directly visible.

\begin{table}[H]
\centering
\caption{City-level geolocation performance (top 10 by validation samples). Acc@500 reports the fraction of samples localized within $500$\,km of ground truth.}
\label{tab:city_performance}
\begin{tabular}{lcccc}
\toprule
\textbf{City} & \textbf{Samples} & \textbf{Mean (km)} & \textbf{Median (km)} & \textbf{Acc@500} \\
\midrule
El~Obeid    & 171 &  291.8 & 203.3 & 0.85 \\
Cairo       & 152 &  404.0 & 198.4 & 0.76 \\
alger       & 121 &  792.5 & 338.4 & 0.62 \\
amman       & 103 &  725.4 & 547.6 & 0.46 \\
Fes         & 101 & 1285.8 & 850.1 & 0.34 \\
Alexandria  &  98 &  182.5 &  97.9 & 0.93 \\
kuwait      &  81 &  850.3 & 596.2 & 0.44 \\
Omdurman    &  80 &  332.9 & 227.9 & 0.90 \\
Annaba      &  80 &  775.8 & 328.1 & 0.63 \\
Ajman       &  73 & 1250.1 & 817.5 & 0.41 \\
\bottomrule
\end{tabular}
\end{table}

\subsection{Dialect Continuum Analysis}

\subsubsection{Mantel Test Results}\label{sec:mantel}

\begin{figure}[htbp]
    \centering
    \includegraphics[width=0.9\textwidth]{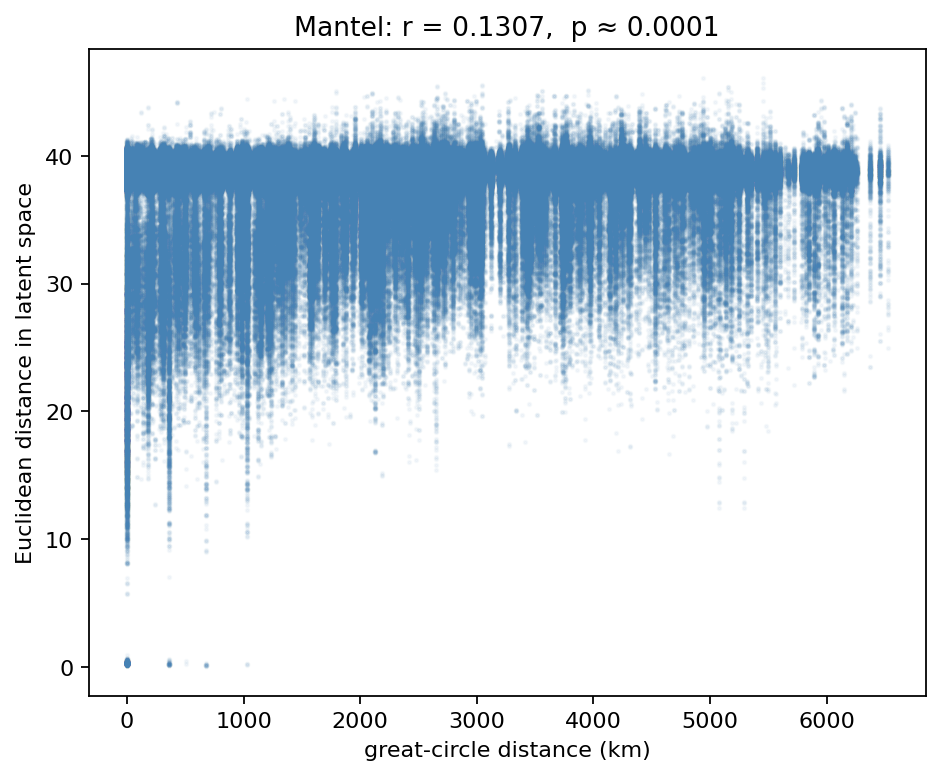}
    \caption{Relationship between geographic and learned linguistic distance for sampled validation pairs. The Mantel correlation $r = 0.131$ ($p \approx 10^{-4}$, $10{,}000$ permutations over $1{,}124{,}250$ pairs) is small but highly significant, evidence that geography is one --- but not the dominant --- structuring axis of the latent space.}
    \label{fig:mantel_correlation}
\end{figure}

We performed a permutation Mantel test correlating Euclidean distance in $\mathbf{h}_{\text{pool}}$ with great-circle distance between ground-truth coordinates, evaluated on $1{,}124{,}250$ validation pairs (drawn from $1{,}500$ samples) and $10{,}000$ permutations. The test yields $r = 0.131$ with $p \approx 1\times 10^{-4}$. The correlation is weak in absolute terms but extremely robust statistically, providing quantitative support for the dialect continuum hypothesis: linguistically similar utterances are systematically more likely to originate from geographically proximate locations. The modest magnitude is consistent with sociolinguistic expectations --- non-geographic factors such as media exposure, education, migration, and individual variation contribute substantial linguistic distance that is decoupled from physical distance. Figure~\ref{fig:mantel_correlation} plots learned linguistic distance against geographic distance for the sampled pairs, illustrating the weak but systematic positive association captured by the test.

\subsubsection{Latent Space Structure}

\begin{figure}[h]
    \centering
    \includegraphics[width=\textwidth]{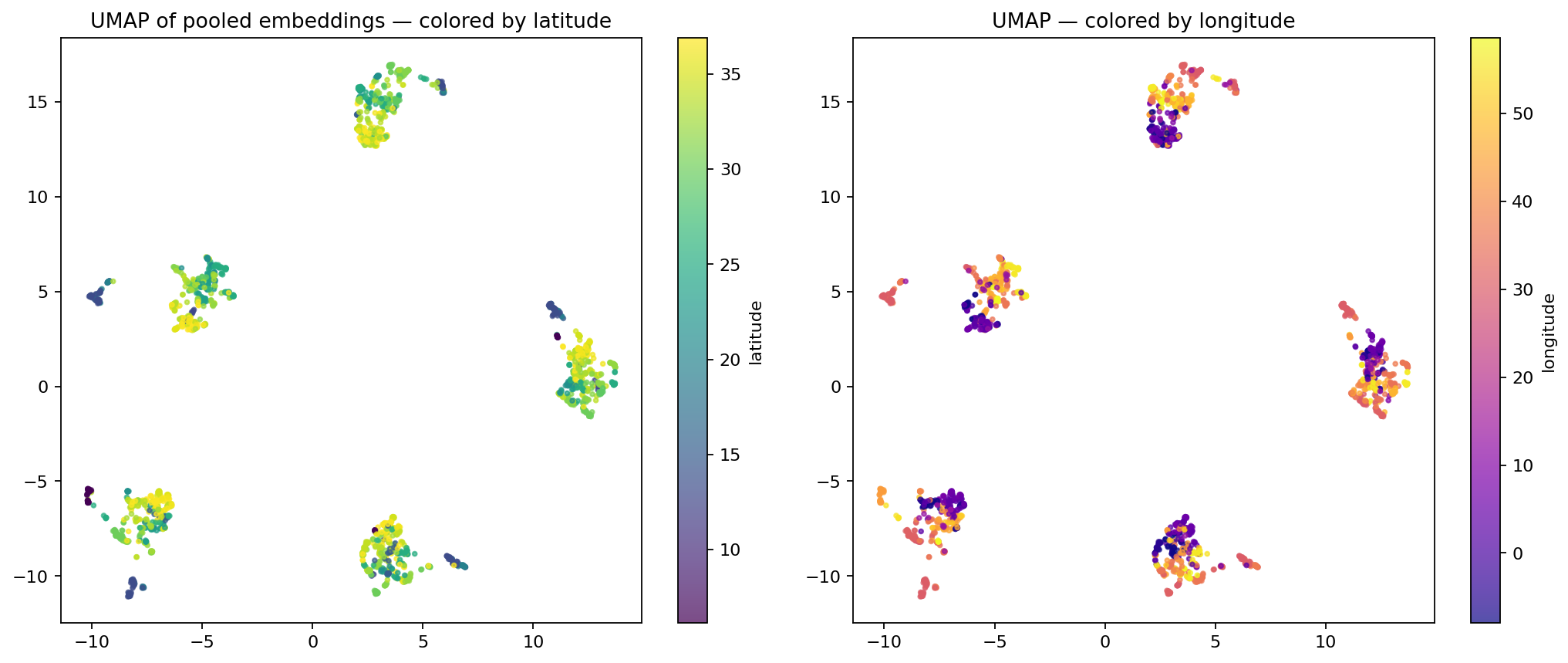}
    \caption{Two-dimensional projection of the learned latent space coloured by ground-truth latitude and longitude. The smooth colour gradients indicate that geographic information is encoded continuously rather than via discrete cluster boundaries.}
    \label{fig:latent_space}
\end{figure}

A two-dimensional t-SNE / UMAP projection of $\mathbf{h}_{\text{pool}}$ coloured by latitude and longitude (Figure~\ref{fig:latent_space}) shows smooth colour gradients rather than discrete clusters separated by sharp boundaries, qualitatively consistent with the Mantel finding. We deliberately do not report numerical ``gradient smoothness'' or ``sharp boundary'' statistics here because we no longer have a calibrated baseline against which their absolute values are meaningful; we treat the projection as a visualization of the structural claim made above rather than as standalone evidence.

\subsection{Border vs. Interior Performance}
\label{sec:border-interior}

\begin{figure}[h]
    \centering
    \includegraphics[width=8.5cm]{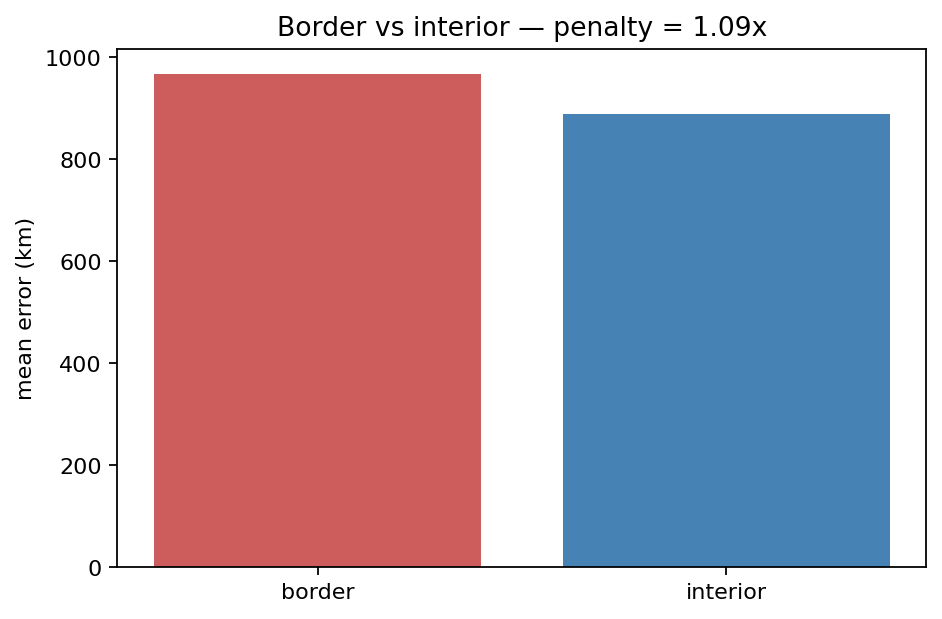}
    \caption{Per-sample geodesic error stratified by border / interior status under the centroid-based heuristic. Distributions overlap heavily for the regression error (border $=1.09 \times$ interior on the mean) but differ markedly for country accuracy ($0.333$ vs.\ $0.702$).}
    \label{fig:border_interior_map}
\end{figure}

Table~\ref{tab:border_interior} reports performance on the centroid-based border / interior split described in Section~\ref{sec:border}. Under this heuristic, $360$ samples are flagged as border-adjacent and $1{,}969$ as interior. Border samples incur a modest but consistent $9\%$ higher mean localization error ($967.7$\,km vs.\ $889.4$\,km, ratio $1.09\times$); the median gap is larger in absolute terms ($688.3$\,km vs.\ $435.6$\,km), reflecting that border-adjacent samples are systematically harder to place precisely. The story is markedly more pronounced at the country head: country accuracy drops from $0.702$ in the interior to $0.333$ at the border --- a $37$-point gap, more than a halving. In other words, the regression head degrades only modestly across boundaries (consistent with the continuum hypothesis), while the discrete country head suffers exactly the kind of catastrophic degradation one would predict for hard-boundary classifiers operating near isoglosses. Figure~\ref{fig:border_interior_map} contrasts the two error distributions, showing the heavy overlap in geodesic error against the sharp divergence in country accuracy.

\begin{table}[H]
\centering
\caption{Border vs.\ interior performance using a $150$\,km nearest-other-country-centroid heuristic. The regression error is $\sim 9\%$ higher at borders, but country-classifier accuracy degrades sharply ($-37$ percentage points).}
\label{tab:border_interior}
\begin{tabular}{lcccc}
\toprule
\textbf{Subset} & \textbf{Samples} & \textbf{Mean (km)} & \textbf{Median (km)} & \textbf{Country Acc.} \\
\midrule
Border    &  360 & 967.7 & 688.3 & 0.333 \\
Interior  & 1969 & 889.4 & 435.6 & 0.702 \\
\bottomrule
\end{tabular}
\end{table}

\subsection{Calibration and Uncertainty Analysis}

\subsubsection{Calibration Quality}

\begin{figure}[H]
    \centering
    \includegraphics[width=14cm]{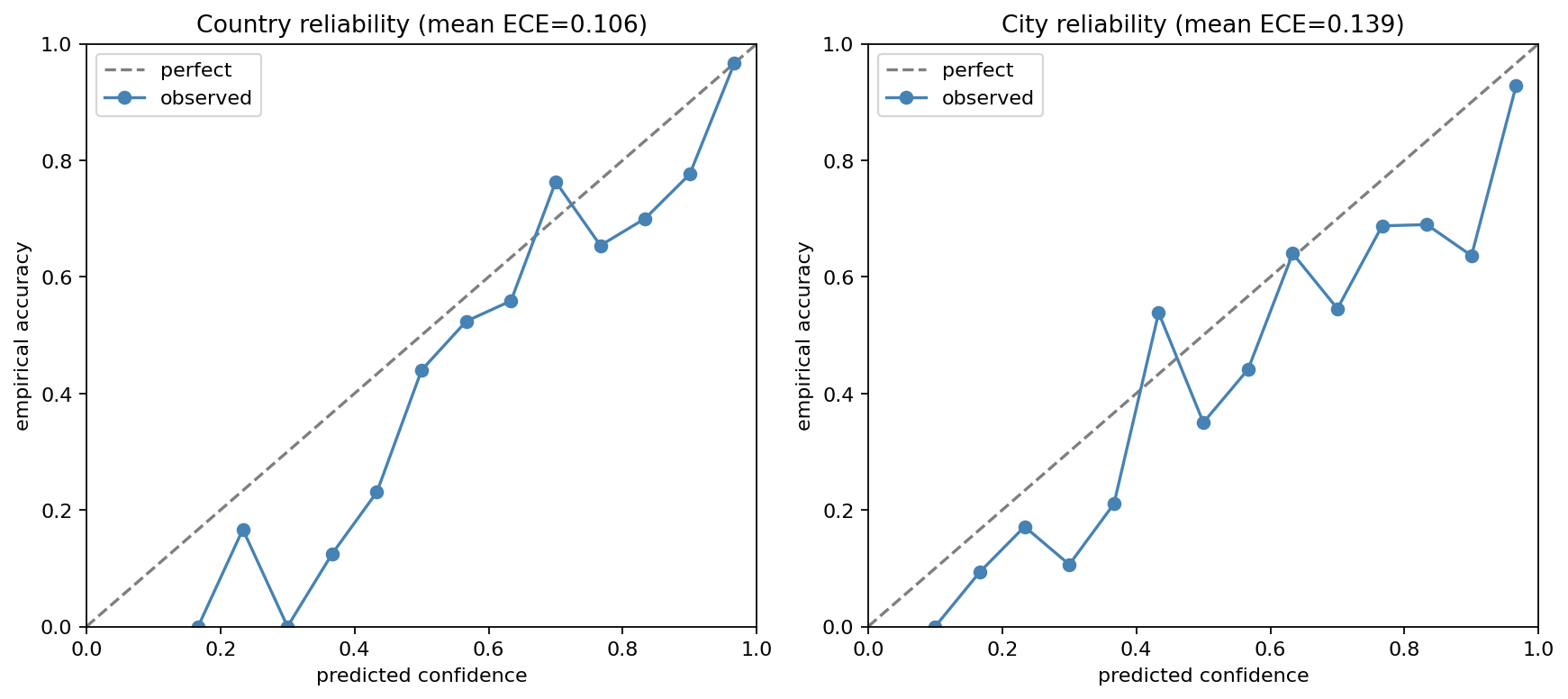}
    \caption{Reliability diagrams for the country and city classification heads. ECE values of $0.106$ (country) and $0.139$ (city), with Brier scores of $0.497$ and $0.716$ respectively, indicate moderate over-confidence in both heads.}
    \label{fig:calibration_curves}
\end{figure}

The two auxiliary heads exhibit moderate calibration error: the country head reaches $\text{ECE} = 0.106$ with Brier score $0.497$, and the city head $\text{ECE} = 0.139$ with Brier score $0.716$. The numbers are stable across folds, suggesting the miscalibration is a property of the model class rather than of any single fold. Both heads tend toward over-confidence, a well-known consequence of cross-entropy training with label smoothing in low-coverage regimes; temperature scaling and isotonic regression are obvious post-hoc remedies that we leave to deployment-time work. The reliability diagrams in Figure~\ref{fig:calibration_curves} make this over-confidence visible, with both curves sitting below the diagonal across most confidence bins.

\subsubsection{Selective Prediction Capability}

Even without explicit calibration, max-softmax confidence is a usable abstention signal. Table~\ref{tab:selective_prediction} reports the operating points obtained by sweeping the per-sample maximum-softmax confidence to retain $80\%$ of validation samples. Conditioning on country-head confidence at threshold $\tau_{\text{country}} = 0.481$ raises country accuracy from $64.5\%$ to $75.3\%$ and pulls the pooled mean geodesic error from $901.5$\,km down to $749.2$\,km (median: $481.2 \to 350.5$\,km) --- a $17\%$ reduction in mean error and $27\%$ reduction in median error for an acceptable abstention rate. Conditioning on city-head confidence ($\tau_{\text{city}} = 0.301$) yields a similar pattern at slightly weaker localization (mean $767.3$\,km, median $365.6$\,km) but pushes city accuracy on the kept subset above $53\%$. 

\begin{table}[H]
\centering
\caption{Selective prediction at $80\%$ coverage ($n_{\text{kept}} = 1863$ out of $2329$). Threshold is chosen on the corresponding head's max-softmax confidence; metrics are pooled across folds.}
\label{tab:selective_prediction}
\begin{tabular}{lccccc}
\toprule
\textbf{Conf.\ head} & \textbf{Threshold} & \textbf{Country Acc.} & \textbf{City Acc.} & \textbf{Mean (km)} & \textbf{Median (km)} \\
\midrule
Country & 0.481 & 0.753 & 0.527 & 749.2 & 350.5 \\
City    & 0.301 & 0.724 & 0.534 & 767.3 & 365.6 \\
\bottomrule
\end{tabular}
\end{table}

\subsection{Unseen-City Zero-Shot Evaluation}
\label{sec:unseen-city}

The city-masking protocol (Section~\ref{sec:cv-protocol}) lets us measure how the model behaves on cities that were never visible during training. Aggregating over the eight masked cities across folds 1--4, mean error on unseen-city samples is $1173.3$\,km ($n = 99$) compared to $889.4$\,km on seen-city samples of the same folds ($n = 2230$). The ratio of $1.32\times$ quantifies the cost of a strict zero-shot evaluation: substantial, but smaller than what one might expect for a system that has no in-distribution view of those cities.

Per-city numbers (Table~\ref{tab:unseen_city}) reveal sharp heterogeneity. Some held-out cities collapse: Tripoli ($2192.8$\,km mean) is mislocated outside Libya entirely, Aleppo ($1506.8$\,km) is dragged toward the Maghreb, and batna's mean ($1463.3$\,km) is dominated by a long tail even though its median ($699.8$\,km) shows that most predictions stay within Algeria. Sharjah ($1345.2$\,km) is consistently confused with other Gulf locations. Others degrade only mildly: tunis ($497.7$\,km mean) and El~Obeid ($597.0$\,km) are essentially as accurate as a typical seen city, presumably because other cities of the same dialect region (alger, Annaba and the Sudanese cities Cairo / Omdurman, respectively) are present in training and span the relevant dialect neighborhood. This heterogeneity is the most informative observation in the paper: zero-shot performance is dominated by how well the \emph{rest of} the training set covers the geographic and dialectal neighborhood of the held-out city, not by any intrinsic difficulty of that city.

\begin{table}[H]
\centering
\caption{Per-city zero-shot evaluation: each city is held out of training in exactly one fold and evaluated on its (still-present) validation samples in that fold. The right-most pair of columns reports the corresponding seen-city baseline from fold~0 (where available), where the same city was not masked.}
\label{tab:unseen_city}
\small
\setlength{\tabcolsep}{4pt}
\begin{tabular}{l c c c c c c}
\toprule
\textbf{Held-out city} & \textbf{Fold} & \textbf{$n_{\text{unseen}}$} & \textbf{Mean (km)} & \textbf{Median (km)} & \textbf{Seen mean (km, fold 0)} & \textbf{$n_{\text{seen, f0}}$} \\
\midrule
Tripoli      & 2 & 12 & 2192.8 & 2340.8 & 1581.6 & 10 \\
Aleppo       & 4 &  6 & 1506.8 & 1303.2 & 1169.5 &  6 \\
Batna        & 1 & 12 & 1463.3 &  699.8 &  233.9 &  9 \\
Sharjah      & 3 & 29 & 1345.2 & 1365.9 & --     &  0 \\
Makkah       & 2 &  1 & 1150.1 & 1150.1 & 1099.3 &  8 \\
Cheikh Taba & 4 &  4 &  669.9 &  713.4 &  759.2 &  2 \\
El~Obeid     & 1 & 30 &  597.0 &  485.3 &  237.5 & 44 \\
Tunis        & 3 &  5 &  497.7 &  482.7 &  487.3 &  1 \\
\midrule
\textbf{Pooled (unseen, $n=99$)}   & -- & 99   & \textbf{1173.3} & -- & -- & -- \\
\textbf{Pooled (seen,   $n=2230$)} & -- & 2230 & \textbf{889.4}  & -- & -- & -- \\
\textbf{Ratio}                     & -- & --   & $\mathbf{1.32\times}$ & -- & -- & -- \\
\bottomrule
\end{tabular}
\end{table}

\subsection{Discussion and Implications}

\subsubsection{Theoretical Implications}

Our results provide qualitative support for the Arabic dialect continuum hypothesis, but with important quantitative caveats. The Mantel test on the latent space is small in magnitude ($r = 0.131$) yet highly significant; t-SNE projections show smooth geographic gradients rather than discrete clusters; and the regression head's mean error increases by only $9\%$ for border-adjacent samples relative to interior ones, even though the discrete country head shows the large $37$-point border degradation expected of any hard-boundary classifier. Taken together these observations are consistent with a continuum, but they do not by themselves rule out a discrete-but-noisy generative process; the present work establishes the leakage-free baseline against which future, more controlled tests of the continuum hypothesis can be run.

\subsubsection{Practical Applications}

Beyond theoretical insights, our approach enables several practical applications. First, selective prediction allows deployment in scenarios requiring high reliability, with the system abstaining when confidence is insufficient. Second, hierarchical predictions (country, city, coordinates) provide graceful degradation and interpretable outputs across multiple granularities. Third, the continuous nature of predictions enables applications requiring precise geographic resolution, such as forensic speaker analysis or targeted content localization.

The model's learned representations could also support downstream tasks. The latent space $\mathbf{h}_{\text{pool}}$ encodes dialectal information that may improve speech recognition, machine translation, or dialect adaptation systems. Transfer learning from our pre-trained encoder could accelerate development of specialized dialect processing tools.

\section{Limitations}\label{sec8}
This work is subject to several limitations related to data quality, modeling assumptions, and evaluation methodology.

\textbf{Data and Annotation Limitations.}
The dataset exhibits strong geographic imbalance, with many cities represented by very few samples and rural dialects largely absent; underrepresented countries (Qatar, Bahrain, Oman) and rural dialects consequently receive much less accurate predictions. This limits generalization to underrepresented regions. Additionally, geographic labels are approximated using city centroids derived from self-reported origins, introducing unavoidable localization error and preventing fine-grained spatial accuracy. Migration histories and dialectal accommodation are not explicitly modeled, further affecting label reliability. Speakers may also be influenced by more than one dialect, for instance expatriates or individuals of mixed heritage, and such multi-dialect influence is not captured by the single-origin labeling scheme. The border/interior split used in our analysis (Section~\ref{sec:border-interior}) further relies on a centroid-distance heuristic rather than true country polygons, adding an additional source of label imprecision.

\textbf{Model Limitations.}
The model produces point coordinate estimates without uncertainty quantification, and auxiliary classifiers are poorly calibrated, limiting reliability in downstream or high-stakes use cases. Calibration quality, while reasonable (ECE $\approx 0.11$ country, $\approx 0.14$ city), is not production-ready and would benefit from temperature scaling or isotonic regression at deployment. Moreover, despite hierarchical outputs, the learned acoustic representations remain largely uninterpretable, restricting linguistic insight beyond aggregate performance. Performance on cities unseen during training (Section~\ref{sec:unseen-city}) also degrades substantially, by $1.32\times$ on average, with very high variance: cities surrounded by other in-vocabulary cities of the same dialect region (e.g., Tunis, El~Obeid) degrade only mildly, while geographically isolated cities (Tripoli, Aleppo, Sharjah in our protocol) degrade severely. Addressing this would require either denser coverage in underrepresented regions or explicit modelling of dialect-region prototypes.

\textbf{Linguistic Limitations.}
Code-switching between dialects and Modern Standard Arabic weakens dialect-specific cues, reducing localization accuracy; more generally, the model relies on dialectal features that are attenuated by code-switching and formal register, making such utterances systematically harder to localize. Within-city variation due to socioeconomic, generational, and stylistic factors is not captured, as annotations assume dialectal homogeneity at the city level.

\textbf{Evaluation Limitations.}
Geodesic distance metrics do not account for dialectal similarity, treating all spatial errors equally regardless of linguistic relevance. Furthermore, the absence of standardized benchmarks for Arabic dialect geolocation limits direct comparison with prior work.

\section{Conclusion}\label{sec9}

We have presented a regression-based approach to Arabic dialect geolocation that models dialectal variation as a continuous geographic space rather than discrete categories. By formulating the task as regression over spherical coordinates and combining frame-level XLS-R-300M and Whisper-large-v3 features with phonotactic descriptors through a Transformer encoder and a learnable attention-pool query, our system avoids the classification--quantization error of city-bin baselines and the planar distortion of latitude--longitude regression.

Under a leakage-free 5-fold GroupKFold protocol grouped by source recording, our model attains a pooled mean error of $901.5$\,km and a median error of $481.2$\,km, with $51.7\%$ of predictions within $500$\,km, $69.5\%$ within $1000$\,km, and $90.9\%$ within $2500$\,km of ground truth. Auxiliary country and city heads reach pooled accuracies of $64.5\%$ and $45.2\%$, respectively, and a confidence-driven selective-prediction policy at $80\%$ coverage further reduces mean error from $901.5$\,km to $749.2$\,km (median: $481.2 \to 350.5$\,km). A permutation Mantel test on the learned latent space ($r = 0.131$, $p \approx 10^{-4}$) provides quantitative, leakage-free support for the Arabic dialect continuum hypothesis.

Methodologically, one of the most important contributions of this paper is the city-masking cross-validation protocol, which lets us measure true zero-shot generalization to held-out cities. Mean error on unseen cities rises to $1173.3$\,km (a $1.32\times$ degradation relative to seen cities) and the spread across cities is dominated by how well the rest of the training set covers the held-out city's dialect region. This is, to our knowledge, the first leakage-free zero-shot evaluation of Arabic dialect geolocation on this corpus.

Future work should focus on direct comparisons against discrete classification baselines under the same leakage-free protocol, polygon-based border analyses, expanded geographic coverage of underrepresented dialects, and post-hoc confidence calibration for deployment.

\begin{acknowledgments}
The authors would like to thank Prince Sultan University for their support.
\end{acknowledgments}

\bibliographystyle{compling}
\bibliography{sample}

\end{document}